\documentclass[11pt]{scrartcl} 
\ifx\pdfminorversion\undefined
\else
\fi

\usepackage{fullpage,epsfig}
\usepackage{comment}
\usepackage{xcolor}
\usepackage{longtable}
\usepackage{multirow,array}
\usepackage{authblk}
\PassOptionsToPackage{hyphens}{url}
\usepackage{hyperref}
\usepackage{color,xspace}
\usepackage{cancel}

\usepackage{enumitem}

\usepackage{amsmath,amsfonts}
\usepackage{algorithmic}
\usepackage{algorithm}
\usepackage{array}
\usepackage{textcomp}
\usepackage{stfloats}
\usepackage{url}
\usepackage{verbatim}
\usepackage{graphicx}
\usepackage{cite}

\usepackage{hyperref}
\usepackage[nameinlink]{cleveref}
\Crefname{figure}{{Fig.}}{{Figs.}}
\Crefname{table}{{Table.}}{{Tables.}}
\Crefname{equation}{{}}{{}}
\Crefname{Algorithm}{}{}
\usepackage[ruled,vlined,linesnumbered,algo2e]{algorithm2e}
\usepackage{booktabs}
\usepackage{xcolor}
\usepackage{lineno}
\usepackage{subfigure}

\newcommand{\remove}[1]{}

\hypersetup{
  colorlinks=true,
  citecolor=blue,
  linkcolor=blue,
  urlcolor=green}
  
\def\BibTeX{{\rm B\kern-.05em{\sc i\kern-.025em b}\kern-.08em
    T\kern-.1667em\lower.7ex\hbox{E}\kern-.125emX}}

\usepackage{amssymb,amsfonts}
\usepackage{wrapfig}
\usepackage{makecell}
\usepackage{tabularx}
\usepackage{diagbox} 
\usepackage{soul}

\usepackage{booktabs}
\usepackage[ruled,vlined,linesnumbered]{algorithm2e}
\Crefname{figure}{{Fig.}}{{Figs.}}
\Crefname{equation}{{}}{{}}
\Crefname{table}{{Table}}{{Tables}}

\title{Hardware-in-the-loop Simulation Testbed for Geomagnetic Navigation}

\author{Songnan Yang\footnote{S. Yang, X. Zhang, and X. Ma are with Xi'an University of Technology, China (yang.son.nan@gmail.com, xhzhang@xaut.edu.cn,  xuehui.yx@gmail.com).} \; Shiliang Zhang\footnote{S. Zhang is with University of Oslo, Norway (shilianz@uio.no).} \; Qianyun Zhang\footnote{Q. Zhang is with University of Toronto, Canada (qianyunzhang\_24@mail.utoronto.ca).\\This work was supported in part by the National Major Scientific Instrument Development Project of China under Grant 62127809; in part by the National Natural Science Foundation of China under Grant 62073258; and in part by the Basic Research in Natural Science and Enterprise Joint Foundation of Shaanxi Province under Grant 2021JLM-58 (Corresponding author: Xiaohui Zhang).} \; Xiaohui Zhang$^\ast$ \; Xuehui Ma$^\ast$}

\begin{document}
\maketitle

\vspace{-0.5cm}
\begin{abstract}
Geomagnetic navigation leverages the ubiquitous Earth's magnetic signals to navigate missions, without dependence on GPS services or pre-stored geographic maps. It has drawn increasing attention and is promising particularly for long-range navigation into unexplored areas. Current geomagnetic navigation studies are still in the early stages with simulations and computational validations, without concrete efforts to develop cost-friendly test platforms that can empower deployment and experimental analysis of the developed approaches. This paper presents a hardware-in-the-loop simulation testbed to support geomagnetic navigation experimentation. Our testbed is dedicated to synthesizing geomagnetic field environment for the navigation. We develop the software in the testbed to simulate the dynamics of the navigation environment, and we build the hardware to generate the physical magnetic field, which follows and aligns with the simulated environment. The testbed aims to provide controllable magnetic field that can be used to experiment with geomagnetic navigation in labs, thus avoiding real and expensive navigation experiments, \textit{e.g.}, in the ocean, for validating navigation prototypes. We build the testbed with off-the-shelf hardware in an unshielded environment to reduce cost. We also develop the field generation control and hardware parameter optimization for quality magnetic field generation. We conduct a detailed performance analysis to show the quality of the field generation by the testbed, and we report the experimental results on performance indicators, including accuracy, uniformity, stability, and convergence of the generated field towards the target geomagnetic environment. 

\end{abstract}


\section{Introduction}
\label{sec:introduction}


Geomagnetic navigation has been extensively investigated as an alternative to GPS-based navigation~\cite{10713176,DBLP:journals/tgrs/QiXXLR23}, especially in long-range missions where the navigation area is unexplored and GPS services or pre-store geographic maps are inaccessible~\cite{DBLP:journals/tim/ZhangZSWZ21,DBLP:journals/corr/abs-2403-08808,bai2024long}. Geomagnetic navigation uses the ubiquitous Earth magnetic field signals to navigate missions~\cite{Chen_2023,Qi_2023}, and recent works have demonstrated its performances close to the navigation supported by reference maps or INS devices~\cite{Goldenberg,Chen_2018}. While theoretical analyses on geomagnetic navigation approaches are sufficient, the developed approaches can be subject to real world constraints. \textit{E.g.}, the magnetic contour matching (MAGCOM) method~\cite{Canciani_2022} and the iterative closest contour point (ICCP) method~\cite{Xiao_2019} - which are considered theoretically highly reliable - cannot be applied in real missions due to the changing magnetic field that is supposed to be static in their method design. Geomagnetic navigation is largely confined in its early stage with modeling and simulations, and the stability of various approaches, \textit{e.g.}, LDUGN method~\cite{Zhang_2021_2)}, the EVO \cite{Zhang_2021} method, and the 2-D method~\cite{Chen_2022}, remains uncertain when it comes to interferences in real-world navigation environments. There is a need of real-world experiments to validate and refine the developed geomagnetic navigation approaches, so as to generate concrete insights for application and deployment. 
Nevertheless, real navigation experiments, \textit{e.g.}, for underwater missions, can be expensive, and it is impractical to conduct massive real experiments with diverse navigation conditions to gain generalized geomagnetic navigation models. Therefore, there is an urgent demand for a low-cost, configurable, and reusable laboratory testbed to evaluate the impact of real geomagnetic conditions on the performance of geomagnetic navigation.

This paper looks into the design of a laboratory testbed facilitating geomagnetic navigation experimentation. Particularly, we examine the synthesis of geomagnetic field environment~\cite{Liu_2023} by a testbed. We consider a dynamic field as the target of the field generation, since the geomagnetic dynamics significantly impacts the geomagnetic navigation performance~\cite{Vald_s_Abreu_2021}. Though synthesizing geomagnetic field seems promising in facilitating navigation experiments in lab, it is challenging to configure and control the magnetic field following reference dynamics and quality indicators. In the following, we review existing approaches in generating magnetic environments and analyze their merits and disadvantages.

Magnetic field can be generated by a coil with well-controlled fed-in current~\cite{Beiranvand_2017}. The current is controlled over time to produce the magnetic field with specific strength and direction~\cite{Abbott_2020}. Helmholtz coils have served as established devices for producing a region of nearly uniform magnetic field~\cite{Jin_2022,Tang_2023}. Amongst different types of coils, square Helmholtz coils can avoid the loss of uniformity of the generated field compared with circular Helmholtz coils~\cite{Zhu_2023}, and square Helmholtz coils can generate more compact magnetic field in a lab environment~\cite{Zhang_2021_3}. Hurtado \textit{et al.}~\cite{Hurtado_Velasco_2016} demonstrated that using square Helmholtz coils can provide superior magnetic field uniformity compared to their circular counterparts, while avoiding the nontrivial coil product alignment. They also unveiled that the magnetic field uniformity - which determines the homogeneousness of the field distribution within a specific area - of the coil depends on its diameter and the spacing between the coils. Cao \textit{et al.}~\cite{Quanliang_Cao_2012} proposed a combination of Maxwell coil and circular Helmholtz coil for magnetic field synthesis, where they generate strong magnetic field with low current that leads to lower costs. Huang \textit{et al.}~\cite{Huang_2020} introduced a square Helmholtz system designed for electromagnetic interference resistance analysis. Their system requires a shielding environment to avoid instability of the generated field caused by environmental noise. Jiang \textit{et al.}~\cite{Jiang_2021} developed a closed-loop control for circular Helmholtz coil to promote magnetic field generation accuracy using model predictive control~\cite{https://doi.org/10.1155/2017/9402684} and super-twisting sliding mode observers. Nevertheless, their control approach demand heavy computation that limits its capacity in generating magnetic field with rapid dynamics. Liu \textit{et al.}~\cite{Liu_2022} designed square Helmholtz coils to generate magnetic field to compensate the geomagnetic anomalies and calibrate the navigation information. They use the coils to generate a stable and static magnetic field to confine environmental disturbances. Nevertheless, the coil's ability to dynamically adjust the magnetic field is not analyzed. Ponikvar \textit{et al.}~\cite{Ponikvar_2023} employed adaptive least mean squares (LMS) to compensate the current fluctuations in square Helmholtz coils for stable and accurate magnetic field generation. However, the convergence of their generated field is compromised by the steady-state errors induced by the fixed step size in LMS. The Aerospace CubeSat project~\cite{Cervettini_2021,Uscategui_2023,Walleshauser_2024} leveraged square Helmholtz coil in designing testbed to support geomagnetic navigation deployment, where they developed geomagnetic navigation testbed using Helmholtz coils in synthesizing magnetic field. Their coils are enclosed in a magnetically shielded room, where the generated magnetic field is free from interference from the external environment. However, the construction of a shielded laboratory room is expensive due to the incorporation of precious materials like mu-metal~\cite{Mateos_2018}. 

Beyond studies on accurate field generation, there are works looking into uniformity of the generated field by Helmholtz coils. Li \textit{et al.}~\cite{Li_2024_2} proposed a parameter design approach for circular Helmholtz coils, where they transfer the coil parameter design problem to a multi-objective optimization problem, and provide optimal parameters for the coil to mitigate field uniformity degradation. However, the experimental performance of the coil with optimized parameters is not discussed to validate the optimization. Lu~\cite{Lu_2022} \textit{et al.} established the relationship between coil parameters and the field uniformity. They verified their approach through finite element method (FEM) simulations, yet their work neglects the field dimension in the uniform region. Lu \textit{et al.}~\cite{Lu2023} proposed an optimization for square Helmholtz coils, where they employed Taylor series expansions for the magnetic field to enhance the uniformity of the target field. They transformed the coil design into a constrained optimization problem, from which they obtained the optimal coil parameters by particle swarm optimization algorithm. Nonetheless, their design is confined within a magnetically shielded barrel, resulting in a limited area of uniformity. Moreover, their work ignores magnetic disturbances from the environment which requires additional current compensation measures. Batista \textit{et al.}~\cite{Batista2019} investigated the design and testing of a voltage-controlled current source (VCCS) for square Helmholtz coils. They demonstrated that VCCSs constructed with low-cost, commercially available components can meet the requirements for uniform field generation in lab. However, their work focused on VCCS design without analysis on target magnetic field generation. Chen \textit{et al.}~\cite{Chen2016} developed a multi-square Helmholtz coil magnetic field generator with an expanded uniform region. Though their method significantly enhances the uniform area of the magnetic field, the field strength fluctuates considerably due to the absence of magnetic interference suppression. Furthermore, their use of multiple coils increases installation cost and complexity.


The analyzed studies above show the dominance of square Helmholtz coil in magnetic field generation. Yet issues exist with Helmholtz coils regarding coil parameter optimization, rapid response, and the uniformity, stability, and convergence of the field generation due to inefficient coil control. Without a proper control, the coil cannot generate the magnetic field pertinent to the requirements in geomagnetic navigation experiments. However, it is nontrivial to design a control that simultaneously coordinate and balance the quality indicators like accuracy, stability, and convergence~\cite{DBLP:journals/tac/MaZLQSH24,10055961,DBLP:conf/eucc/MaCZLQS24,yu2024adaptive,ma2022adaptive}. Furthermore, though there exist one testbed that has been developed in supporting geomagnetic navigation, it is costly to build such a testbed with shielding environment thus jeopardizing its widely adoption in facilitating experimentation.

In this paper, we aim to address those challenges by designing a testbed with efficient coil parameter optimization and field generation control for magnetic field generation, in an unshielded lab environment. To this end, we build a hardware-in-the-loop simulation testbed that empowers geomagnetic navigation experiments in lab. We design the testbed in a way that it can synthesize both stable and rapidly varying magnetic fields, the latter of which commonly exists in geomagnetic anomalies. Particularly, in the software part of our testbed, we simulate a geomagnetic field induced by a virtual square Helmholtz coil using finite element method (FEM)~\cite{DBLP:journals/tim/TangDJF23}. We determine the coil parameters fed to FEM by designing a parameter optimization approach, which guarantees that the coil can maximize a uniform field area. The hardware part of our testbed is a physical counterpart of the virtual Helmholtz coil. To adjust the magnetic field generated by the physical coil and make it aligned with the virtual field by FEM, we develop a convex combination control approach for the Helmholtz coil. We adopt a convex combination approach to concurrently coordinate the accuracy, stability, and convergence for the field generation. The developed control can adjust the current fed to the coil in generating a magnetic field following reference dynamics, while compensating the interferences from the external environment without an expensive shielded environment. We build our testbed with off-the-shelf hardware like micro controller units and magnetometer sensors, and we provide mathematical proofs for the stability and convergence of the generated field towards the target value. We conduct comprehensive experiments and detailed analyses to demonstrate our testbed regarding the quality of generated field, and we compare our designed coil control with existing Helmholtz coil control approaches. Note that while our ultimate goal is the experimentation of geomagnetic navigation with a testbed in lab, we in this paper demonstrate the design, deployment, and validation of our testbed on magnetic field generation, along with the developed coil optimization and control that underpin a quality field generation. We summarize the contribution of this work as follows.

\begin{enumerate}
    \item We design and build a cost-friendly testbed with off-the-shelf hardware in an unshielded laboratory environment to support repeatable geomagnetic navigation experiments, thus avoiding expensive real navigation missions in validating navigation prototypes. 
    

    \item We build our testbed in a hardware-in-the-loop architecture, where the software is a digital twin of the hardware part that work together for magnetic field synthesis. The hardware and software are aligned with each other to guarantee that a target virtual navigation environment in the simulation is generated physically and precisely by the hardware.
    

    \item We design the square Helmholtz coil parameter optimization approach to fit the coil design for different carriers, \textit{e.g.}, with different requirements on the magnetic field size. Furthermore, we propose a convex combination coil control method for quality field generation. The proposed control provides a proofed convergence of the synthetic field towards the target value that ensures rapid response, stability, and accuracy of the magnetic field generation.
    
\end{enumerate}

The rest of this work is organized as follows. \Cref{sec:Hardware Design and Implementation} explains the design and implementation of square Helmholtz coils under Biot-Savart law, along with the coil parameter optimization. \Cref{sec:Convex Combination Coil Control Method} introduces the convex combination coil control method for our testbed and provides proof of its convergence and stability. \Cref{sec:Hils Performance Validation Methodology} validates the magentic field generation by our testbed and examines the performance of the convex combination coil control method. We conclude this work in \Cref{sec:conclusion}.

\begin{figure*}[tph]
\centerline{\includegraphics[width=0.9\textwidth]{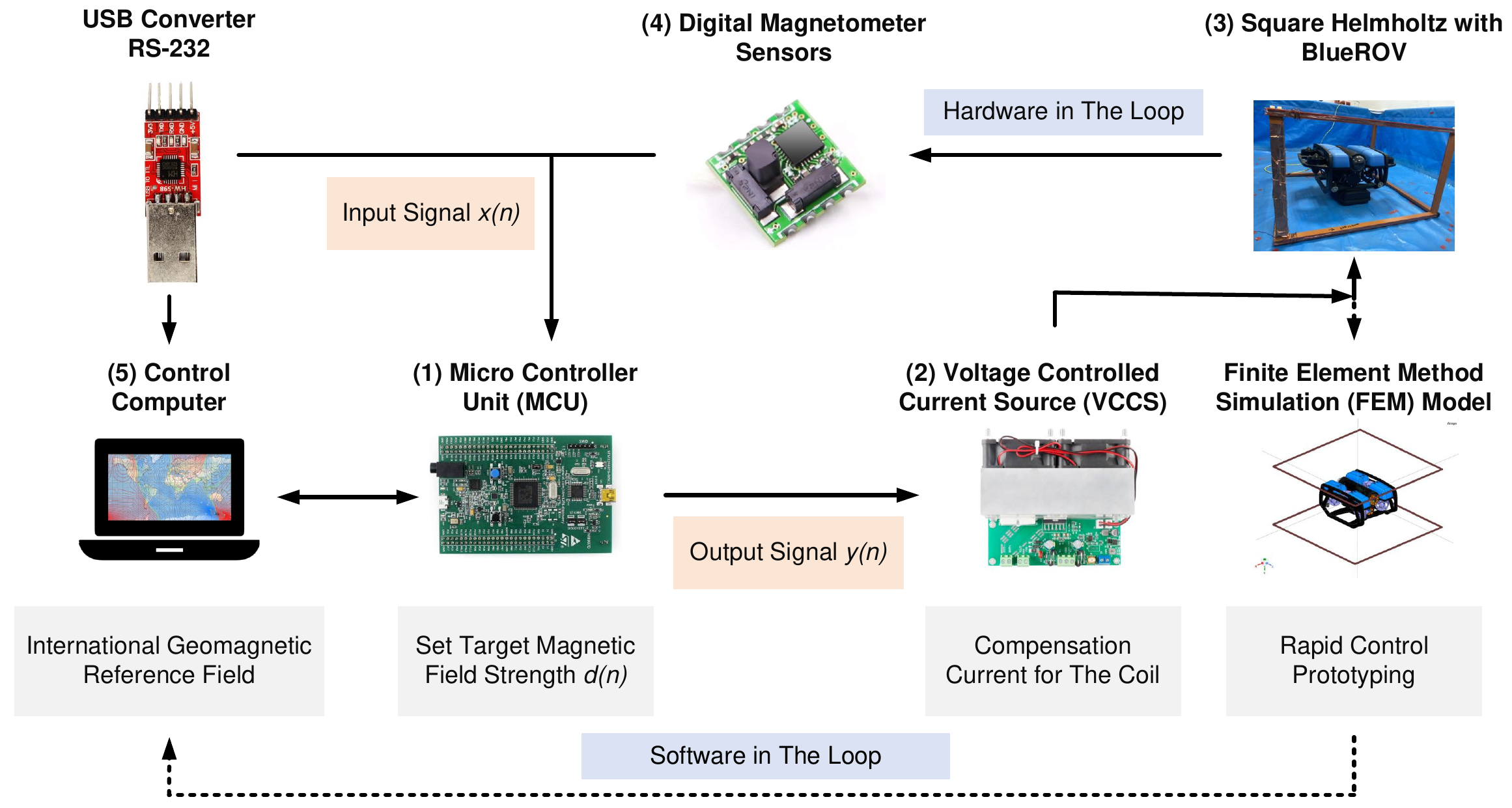}}
\caption{A system architecture of the proposed Hils testbed.}
\label{fig:2}
\end{figure*}

\section{Hardware design, implementation, and parameter optimization}
\label{sec:Hardware Design and Implementation}
We present the architecture of the designed Hardware-in-the-loop simulation (Hils) testbed in~\Cref{fig:2}. The testbed contains five main components: 
(i) a microcontroller unit (MCU) equipped with coil control that transmits and translates the control signals for the adjustment of field generation (ii) a voltage-controlled current source (VCCS) that generates the corresponding current to regulate the magnetic field strength (iii) a square Helmholtz coil to generate a uniform magnetic field, the uniformity of which is crucial for testing geomagnetic navigation performance (iv) digital magnetic sensors that measure the actual magnetic field strength and feed the measurement to the MCU to feedback the error (v) a control computer that generate the virtual navigation environment with target magnetic field, and communicates with the MCU for field synthesis. Note that all the components displayed in~\Cref{fig:2} are off-the-shelf and are mature products that meet the metrics like accuracy and resolution required in the testbed design. 

Beyond the hardware, we also have a digital twin~\cite{10698570} of the physical Helmholtz coil that is connected to the simulated navigation environment by the control computer. The virtual Helmholtz coil is simulated by finite element method (FEM), and is aligned with the physical coil in magnetic field generation. The FEM model simulates the ideal field excited by a well-controlled current under the configuration of the physical coil, while the physical Helmholtz coil requires the control for quality field generation and compensating the interferences from its surrounding environment. Below we explain how the magnetic field can be generated following Biot-savart law, and we demonstrate how to determine the coil parameters that fits a carrier with its specifications in geomagnetic navigation.

\subsection{Magnetic field generated by a coil}
\label{sec:2.1}
\begin{figure}[htbp]
\centering
\subfigure[]
{
    \begin{minipage}[b]{0.4\linewidth}
        \centering
        \includegraphics[width=\textwidth]{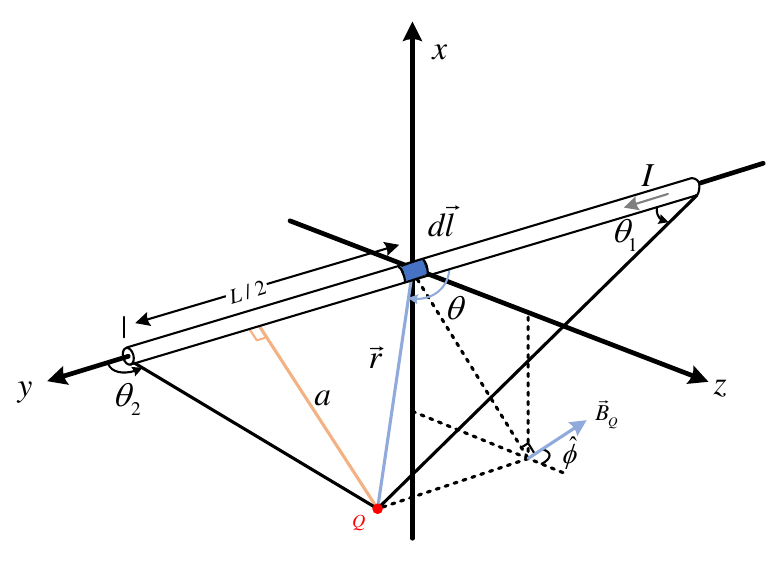}
    \end{minipage}
    \label{fig:3a}
}
\subfigure[]
{
 	\begin{minipage}[b]{0.4\linewidth}
        \centering
        \includegraphics[width=\textwidth]{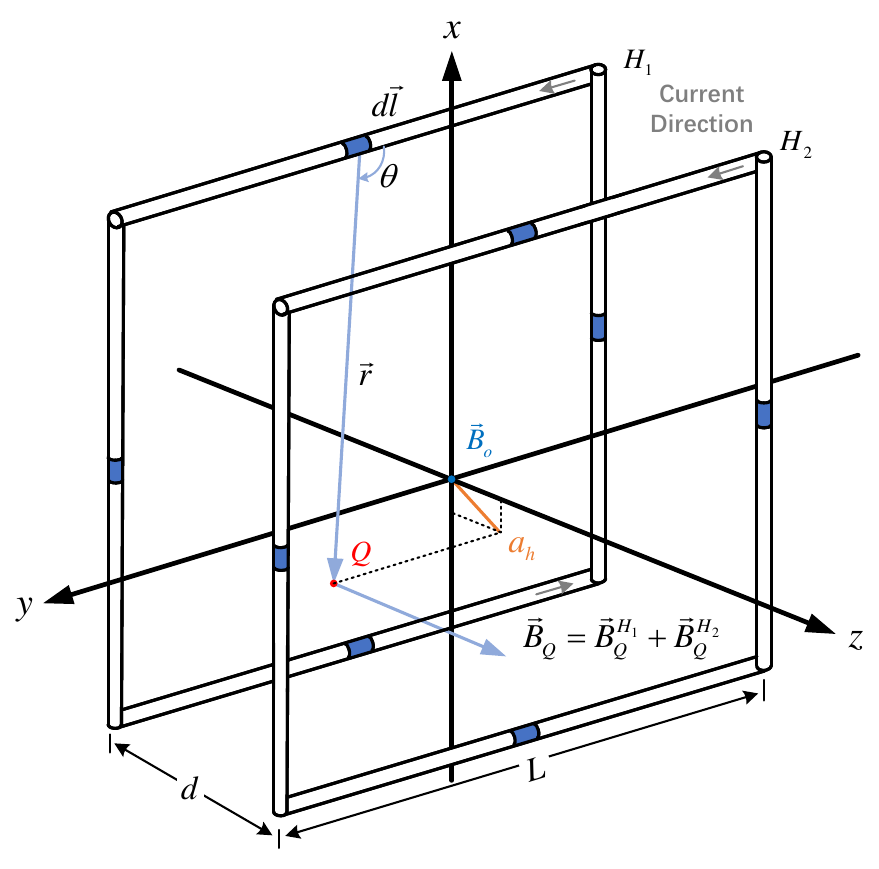}
    \end{minipage}
    \label{fig:3b}
}
\caption{The schematic diagram for magnetic field strength calculation in the Hils testbed.~(a) Field strength calculation at point Q for the magnetic field generated by the current element parallel to the \textit{y}-axis (b) field strength calculation at point Q for a square Helmholtz coil.}
\label{fig:3}
\end{figure}

The current flowing through a conductor can induce a magnetic field near the conductor, following the Bio-Savart law. \Cref{fig:3a} depicts the strength and direction of the induced magnetic field \(d{\boldsymbol{\vec B}}_Q\) at \(Q(x,y,z)\) in space in a conductor under the Bio-Savart law, which is calculated by
\begin{align}
d{\boldsymbol{\vec B}}_Q = N\frac{{{\mu _0}Id\boldsymbol{\vec l}}}{{4\pi |\boldsymbol{\vec r}{|^3}}} \times \boldsymbol{\vec r},
\label{eq:1}
\end{align}
where \(Id\boldsymbol{\vec l}\) denotes the current, \({\mu _0} = 4\pi  \times {10^{ - 7}}~{{T}} \cdot {{m/A}}\) represents the permeability of free space, \(\boldsymbol{\vec r}\) is the vector pointing from the current element \(Id\boldsymbol{\vec{l}}\) to point \(Q\), \(N\) is the number of turns in the coil and in the case in \Cref{fig:3a} \(N=1\). below we illustrate how to calculate the magnetic strength around the coil.


We consider a part of the square Helmholtz coil shown in \Cref{fig:3a}, where \(d\boldsymbol{\vec l} = dy\boldsymbol{\hat y}\) denotes the length of the conductor along the \textit{y}-axis, \(\vec r = a\boldsymbol{\hat a} + y\boldsymbol{\hat y}\) is the vector pointing to point \(Q\), \(\boldsymbol{\hat y}\) and \(\boldsymbol{\hat a}\) represent the unit vectors along the \textit{y}-axis and \textit{a}-axis, respectively. The \textit{a}-axis is orthogonal to the current \(Id\boldsymbol{\vec l}\), and \(a\) denotes the distance from the conductor to point \(Q\). Applying the cross-product rule~\cite{Hurtado_Velasco_2016} yields:
\begin{align}
d\boldsymbol{\vec l} \times \boldsymbol{\vec r} = dy\boldsymbol{\hat y} \times (a\boldsymbol{\hat a} + y\boldsymbol{\hat y}) = ady(\boldsymbol{\hat y} \times \boldsymbol{\hat a}) = ady\boldsymbol{\hat \phi},
\label{eq:2}   
\end{align}
where \(\boldsymbol{\hat \phi} \) represents the unit vector that is perpendicular to the \textit{y}-\textit{a} plane. Substituting\Cref{eq:2} to\Cref{eq:1} yields:
\begin{align}
d{\boldsymbol{\vec B}}_Q = N\frac{{{\mu _0}I}}{{4\pi }}\frac{{ady\boldsymbol{\hat \phi}}}{{{{\left( {\sqrt {{a^2} + {y^2}} } \right)}^3}}}.
\label{eq:3}   
\end{align}

Note that we have \(y = a\cot (\theta )\) from \Cref{fig:3a}. By substituting \(y = a\cot (\theta )\) to\Cref{eq:3},  we have
\begin{align}
d{\boldsymbol{\vec B}}_Q =  - N\frac{{{\mu _0}I}}{{4\pi }}\frac{{a( - a{{\csc }^2}(\theta )d\theta )\boldsymbol{\hat \phi}}}{{{{\left( {\sqrt {{a^2} + {a^2}{{\cot }^2}(\theta )} } \right)}^3}}}.
\label{eq:4}   
\end{align}

Through the integration of \(d{\boldsymbol{\vec B}}_Q\) from \(\theta\) from \(\theta_1\) to \(\theta_2\), the total magnetic field \({\boldsymbol{\vec B}}_Q\) at point \(Q\) can be obtained by:
\begin{align}
{\boldsymbol{\vec B}}_Q =  - N\frac{{{\mu _0}I}}{{4\pi a}}\hat \phi \int_{{\theta _1}}^{{\theta _2}} {\sin \theta d\theta }  = N\frac{{{\mu _0}I}}{{4\pi a}}(\cos {\theta _2} + \cos {\theta _1})\boldsymbol{\hat \phi}.
\label{eq:5}   
\end{align}

\subsection{Magnetic field generated by Helmholtz coil and coil parameter optimization}
\label{sec:2.3}
\Cref{fig:3b} shows the structure of a Helmholtz coil, which is made up of basic elements as shown in \Cref{fig:3a}. Assume that the two square coils in \Cref{fig:3b} are perfectly parallel with a distance \(d\). We segment the coils are into two components for the magnetic field calculation, namely the horizontal coils parallel to the \textit{y}-axis and vertical coils parallel to the \textit{x}-axis. The magnetic field strength at a point \(Q\) results from the cumulative effect of the fields generated by both the horizontal and vertical coils. Below are the detailed calculations. 

We first look into the element \(Id\boldsymbol{\vec l}\) in coil \(H1\) that is parallel to the \textit{y}-axis, as shown in \Cref{fig:3b}. We define \(a_h\) as a projection distance of \(Q\) on the \textit{x}-\textit{z} plane, obtained by
\begin{align}
{a_h} = \sqrt {{{(x - \frac{L}{2})}^2} + {{(z - \frac{d}{2})}^2}}.
\label{eq:6}   
\end{align}
We then define the unit vector \({\boldsymbol{\hat \phi}_h}\) for the horizontal plane as
\begin{align}
{\boldsymbol{\hat \phi}_h} = \cos {\phi _h}\boldsymbol{\hat z} + \sin {\phi _h}\boldsymbol{\hat x},
\label{eq:7}   
\end{align}
where \(\cos {\phi _h} = (x - \frac{L}{2})\slash {a_h}\) and \(\sin {\phi _h} = (z - \frac{d}{2})\slash {a_h}\). 

Similar to ~\Cref{eq:5}, we express \(cos \theta_1\) and \(cos \theta_2\) for the horizontal plane by
\begin{align}
\cos {\theta _{1h}} = \frac{{\frac{L}{2} + y}}{{\sqrt {{{(\frac{L}{2} + y)}^2} + a_h^2} }},
\label{eq:8}   
\end{align}
\begin{align}
\cos {\theta _{2h}} = \frac{{\frac{L}{2} - y}}{{\sqrt {{{(\frac{L}{2} - y)}^2} + a_h^2} }}.
\label{eq:9}   
\end{align}

Substitute~\Cref{eq:8} and~\Cref{eq:9} into~\Cref{eq:5}, and we gain the strength of the magnetic field of the horizontal plane as
\begin{equation}
\begin{aligned}
\boldsymbol{\vec B}_{Qh}^{{H_1}} & {=} N\frac{{{\mu _0}I}}{{4\pi {a_h}}}(\cos {\theta _{2h}} + \cos {\theta _{1h}}){{\boldsymbol{\hat \phi}}_h} \\
&{=} N\frac{{{\mu _0}I}}{{4\pi {a_h}}}(\cos {\theta _{2h}} + \cos {\theta _{1h}})(\cos {\phi _h}\boldsymbol{\hat z} + \sin {\phi _h}\boldsymbol{\hat x}) \\
&{=} N\frac{{{\mu _0}I}}{{4\pi {a_h}}}(\cos {\theta _{2h}} + \cos {\theta _{1h}})\left(\frac{{x - \frac{L}{2}}}{{{a_h}}}\boldsymbol{\hat z} + \frac{{z - \frac{d}{2}}}{{{a_h}}}\boldsymbol{\hat x}\right).
\label{eq:10}
\end{aligned}
\end{equation}

In the same way, we can then obtain the magnetic field strength generated by the vertical part of the coils parallel to the \textit{x}-axis. We define \(a_v\) as a projection distance between the point \(Q\) and the \textit{y}-\textit{z} plane as
\begin{align}
{a_v} = \sqrt {{{(y - \frac{L}{2})}^2} + {{(z - \frac{d}{2})}^2}},
\label{eq:11}   
\end{align}
where the unit vector \({\hat \phi _v}\) is defined by:
\begin{align}
{\hat \phi _v} = \cos {\phi _v}\hat z + \sin {\phi _v}\hat y,
\label{eq:12}   
\end{align}
where \(\cos {\phi _v} = (y - \frac{L}{2})\slash {a_v}\) and \(\sin {\phi _v} = (z - \frac{d}{2})\slash {a_v}\).

Similar to \Cref{eq:5}, we define \(cos \theta_1\) and \(cos \theta_2\) for the vertical plane as
\begin{align}
\cos {\theta _{1v}} = \frac{{\frac{L}{2} + x}}{{\sqrt {{{(\frac{L}{2} + x)}^2} + a_v^2} }},
\label{eq:13}
\end{align}
\begin{align}
\cos {\theta _{2v}} = \frac{{\frac{L}{2} - x}}{{\sqrt {{{(\frac{L}{2} - x)}^2} + a_v^2} }}.
\label{eq:14}
\end{align}

Substitute~\Cref{eq:13} and~\Cref{eq:14} into~\Cref{eq:5}, and we can obtain magnetic strength for the vertical plane as
\begin{align}
\boldsymbol{\vec B}_{Qv}^{{H_1}} = N\frac{{{\mu _0}I}}{{4\pi {a_v}}}(\cos {\theta _{2v}} + \cos {\theta _{1v}})(\frac{{z - \frac{d}{2}}}{{{a_v}}}\boldsymbol{\hat y} + \frac{{y - \frac{L}{2}}}{{{a_v}}}\boldsymbol{\hat z}).
\label{eq:15}
\end{align}

With the magnetic field strength of the horizontal and vertical plane obtained for the coil \(H_1\), we can gain the field strength \(\boldsymbol{\vec B}_Q\) at the point \(Q(x,y,z)\) as
\begin{align}
\boldsymbol{\vec B}_Q^{{H_1}} = \boldsymbol{\vec B}_{Qh}^{{H_1}} + \boldsymbol{\vec B}_{Qv}^{{H_1}},
\label{eq:16}
\end{align}
and we can also obtain the field strength generated by the coil \(H_2\) at \(Q(x,y,z)\) as \(\boldsymbol{\vec B}_Q^{{H_2}}\) in the same way.

Experimentation with the magnetic field requires the uniformity of the generated field within a certain area, \textit{e.g.}, an area that can cover the size of a carrier like an autonomous underwater vehicle. Below we show how to optimize the coefficients of the Helmholtz coil such that the uniformity of the generated field at the centre of the coil can be maximized. 

The uniformity of the magnetic field at the point \(Q\) can be calculated as:
\begin{align}
H[\% ] = \frac{{{{\boldsymbol{\vec B}}_Q} - {\boldsymbol{\vec B}_o}}}{{{\boldsymbol{\vec B}_o}}} \times 100,
\label{eq:17}
\end{align}
where \(\boldsymbol{\vec B}_o\) is the magnetic field strength at the origin \((0,0,0)\) and the field strength at point \(Q\) under the joint effect of coils \(H_1\) and \(H_2\) shown in~\Cref{fig:3b} can be calculated as: 
\begin{align}
{\boldsymbol{\vec B}_Q} = \boldsymbol{\vec B}_Q^{{H_1}} + \boldsymbol{\vec B}_Q^{{H_2}}
\label{eq:53}
\end{align}

Note that though the magnetic field and its uniformity are of three dimensions (x-, y-, and z-axis directions), the field strength near the center of the Helmholtz coil is almost constant in the x- and y-axis direction, while the field strength in the z-axis direction varies in proportion to the current flowing through the coils~\cite{Li_2021, Wang_2023, Leni_ek_2022}. Therefore, we only analyze the field strength and uniformity in the z-axis direction. This analysis is fully applicable to the case of three-dimensional uniformity where additional coils are placed in the x- and y-axis direction.

According to\Cref{eq:10},\Cref{eq:15} and\Cref{eq:16} the magnetic field strength at the point \(Q\) in the \textit{z}-axis direction generate by the coil \(H_1\) can be calculated as:
\begin{align}
\boldsymbol{\vec B}_Q^{{H_1}}(z) {=} \frac{{2N{\mu _0}I}}{\pi }\!\left(\! {\frac{{{{(\frac{L}{2})}^2}}}{{({{(\frac{L}{2})}^2} \!{+}\! {{(z {-} \frac{d}{2})}^2})\!\sqrt {2{{(\frac{L}{2})}^2} \!{+}\! {{\left( {z {-} \frac{d}{2}} \right)}^2}} }}} \!\right)\!\boldsymbol{\hat z}.
\label{eq:18}
\end{align}
Similarly, the field strength at point \(Q\) in z-axis direction generated by the coil \(H_2\) can be obtained by
\begin{align}
\boldsymbol{\vec B}_Q^{{H_2}}(z) {=} \frac{{2N{\mu _0}I}}{\pi }\!\left(\! {\frac{{{{(\frac{L}{2})}^2}}}{{({{(\frac{L}{2})}^2} \!{+}\! {{(z {+} \frac{d}{2})}^2})\!\sqrt {2{{(\frac{L}{2})}^2} \!{+}\! {{\left( {z {+} \frac{d}{2}} \right)}^2}} }}} \!\right)\!\boldsymbol{\hat z}.
\label{eq:19}
\end{align}
then we conduct the Taylor expansion of \(\boldsymbol{\vec B}_Q (z)\) by\Cref{eq:53} at \(z=0\) and it yields
\begin{align}
\boldsymbol{\vec B}_Q\!(\!0\!) {\approx} \boldsymbol{\vec B}_Q\!(\!0\!) {+} z' \!\boldsymbol{\vec B}_Q'\!(\!0\!) {+} z^2 \frac{\boldsymbol{\vec B}_Q''(\!0\!)}{2!} {+} z^3 \frac{\boldsymbol{\vec B}_Q^{(3)}(\!0\!)}{3!} {+} z^4 \frac{\boldsymbol{\vec B}_Q^{(4)}(\!0\!)}{4!}.
\label{eq:20}
\end{align}
Since \(\boldsymbol{\vec B}_Q(z) = \boldsymbol{\vec B}_Q(-z)\) and \(\boldsymbol{\vec B}_Q(z)\) is an odd function, it holds that \(z\boldsymbol{\vec B}(0) = {z^3}\boldsymbol{\vec B}_Q^{(3)}(0)/3!\). We assume that \(L = nd\) in~\Cref{eq:20}, and the second derivative for \(\boldsymbol{\vec B}_Q''(0)\) can be obtained by
\begin{align}
\boldsymbol{\vec B}_Q^{''}\!(0) {= }\frac{{64IN{\mu _0}{n^2}\left( { - 5{n^6}\!{+}\!11{n^4}\!{+}\!18{n^2}\!{+}\!6} \right)}}{{\pi\!{d^2}\!\sqrt {\!\!{d^2} \!\cdot\! \left( {2{n^2}\!{+}\!1} \right)}\! \left(\!{4{n^{10}}\!{+}\!16{n^8} {+} 25{n^6}\!{+}\! 19{n^4}\!{+}\!7{n^2}\!{+}\! 1}\!\right)}}\!\boldsymbol{\hat z},
\label{eq:21}
\end{align}
and the optimized \(n\) can be obtained by solving \(\boldsymbol{\vec B}_Q''(0) = 0\), that is
\begin{align}
- 5{n^6} + 11{n^4} + 18{n^2} + 6 = 0.
\label{eq:22}
\end{align}

Based on~\Cref{eq:22}, the uniformity of the magnetic field for a square Helmholtz coil is optimal when \(n=1.8365\). With the optimal value, we can determine the optimal distance between the coils in the Helmholtz coil that leads to maximized uniformity near the center of the coil. 

\section{Convex combination coil control}
\label{sec:Convex Combination Coil Control Method}
To ensure that the magnetic field at the center of the Helmholtz coil can accurately and rapidly follow a target magnetic field, this section develops a coil control to adjust the current flowing through the coil. The designed control is capable to compensate the current in real-time to mitigate the interference from the surrounding environment. Furthermore, we provide the convergence and stability proofs for the magnetic field generation under the designed coil control. 

\subsection{Coil control method description}
\label{sec:3.1}
\begin{figure}[htbp]
\centerline{\includegraphics[width=0.8\columnwidth]{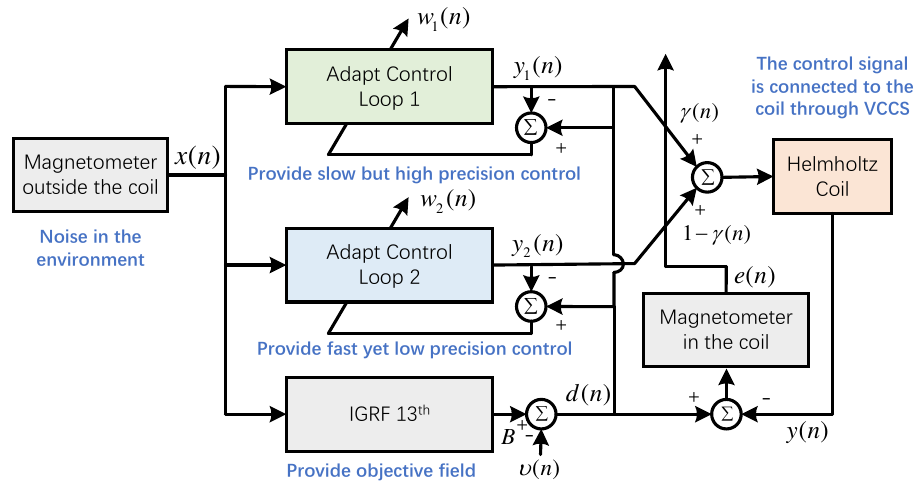}}
\caption{The schematic diagram of the convex combination coil control method. The International Geomagnetic Reference Field (IGRF 13th) block in the figure provides the reference geomagnetic field for a specific location on the Earth.}
\label{fig:4}
\end{figure}

We adopt a convex combination approach~\cite{Arenas_Garcia_2006} to coordinate multiple control objectives concurrently. We design a convex combination coil control with two control loops shown in~\Cref{fig:4}. Control loop 1 in the figure adjust the control signal slowly to achieve a high accuracy and stability of the generated field, while control loop 2 compute control signals faster and adaptively to ensure rapid convergence and response of the generated field toward the target field. The proposed coil control uses adaptive weight transfer techniques to balance multiple objectives for quality magnetic field generation. Particularly, the proposed control aims to achieve a balance between convergence, stability, and accuracy of the generated field for both slow and high precision control by control loop 1 and fast yet low precision control by control loop 2, without increasing computational complexity. 
Below we explain how the designed control works in details.

The proposed control includes six main processes shown below.

\subsubsection{Control signal output}
We use \(\textbf{x}(n)\) to represent the magnetic field strength measured by the magnetometer outside the coil. \(\textbf{x}(n)\) acts as the environmental disturbance for the field generation by the coil. \(\textbf{d}(n)\) is the target field strength provided by the IGRF 13th~\cite{Alken_2021} model at a specific location, \(\textbf{w}_1\) and \(\textbf{w}_2\) are weight vectors representing the controller in the control loop 1 and 2, respectively. 
\(\textbf{y}(n)\) is the control signal that combines the outputs \(\textbf{y}_1 (n)\) and \(\textbf{y}_2 (n)\) from the two controllers through a convex combination. \(\textbf{y}(n)\) is used to adjust the current flowing through the Helmholtz coil. The control signals from the two controllers are calcualted by
\begin{align}
{\textbf{y}_1}(n) = {\textbf{w}_1}^T(n)\textbf{x}(n) + \textbf{B},
\label{eq:24}
\end{align}
\begin{align}
{\textbf{y}_2}(n) = {\textbf{w}_2}^T(n)\textbf{x}(n) + \textbf{B},
\label{eq:25}
\end{align}
\begin{align}
\textbf{B} = k\textbf{x}(n) + b,
\label{eq:26}
\end{align}
\begin{align}
\textbf{y}(n) = \gamma (n)\textbf{w}_1^T(n)\textbf{x}(n) + (1 - \gamma (n))\textbf{w}_2^T(n)\textbf{x}(n) + \textbf{B},
\label{eq:27}
\end{align}
\begin{align}
\textbf{d}(n) = \textbf{B} + \upsilon (n) = {\textbf{w}_o}^T\textbf{x}(n) + \boldsymbol{\varepsilon} (n),
\label{eq:28}
\end{align}
where \(\textbf{B}\) is the target value of the magnetic field, the input is a linear relationship with \(k\) and \(b\) denoting the magnetic field fitting coefficients, \(\upsilon(n)\) represents the noise from the sensors and surrounding environment, \(\boldsymbol{\varepsilon} (n)\) aggregates the noises from all sources. Both \(\upsilon(n)\) and \(\boldsymbol{\varepsilon} (n)\) are assumed to be zero-mean Gaussian white noise.

\subsubsection{Error estimation}
The errors between the outputs of the two controllers and their corresponding target values are calculated by
\begin{align}
{\textbf{e}_1}(n) = \textbf{d}(n) - {\textbf{y}_1}(n),
\label{eq:29}
\end{align}
\begin{align}
{\textbf{e}_2}(n) = \textbf{d}(n) - {\textbf{y}_2}(n),
\label{eq:30}
\end{align}
\begin{align}
\textbf{e}(n) = \textbf{d}(n) - \textbf{y}(n) = \gamma (n){\textbf{e}_1}(n) + (1 - \gamma (n)){\textbf{e}_2}(n),
\label{eq:31}
\end{align}
where \(\textbf{e}(n)\) represents the total error for sensor measurements, \(\gamma(n)\) is the coupling coefficient and \(0 < \gamma (n) < 1\). 

The key of the convex combination method lies in harmonizing the performance of the two controllers. We obtain this harmonization via dynamically modifying the interaction between the two controllers by renewing the update factor \(b(n)\) and the coupling coefficient \(\gamma(n)\). In this paper, we define \(\gamma(n)\) as logistic function to ensure a smooth and bounded transition. Its monotonicity and convexity make it suitable for use as an expression for the coupling coefficient~\cite{Agrawal_2021}, which can be update as follows
\begin{align}
\gamma (n) = \frac{1}{{1 + {e^{ - b(n)}}}}.
\label{eq:32}
\end{align}

\subsubsection{Learning rate update}
We employ both nonlinear and linear learning rates to enable the two controllers to approximate their target concurrently. This facilitates the balance between accurate approximation and rapid convergence of the generated field toward the target value. We explain the learning rate update below
\begin{align}
{\mu _1}(n){=} \beta (\frac{1}{{1{+}\exp ( - \alpha \left| {{\textbf{e}_1}(n){\textbf{e}_1}(n{-}1)} \right| + \sigma \left| {{\textbf{e}_1}(n)} \right|)}}{-}0.5),
\label{eq:33}
\end{align}
\begin{align}
{\mu _2}(n) = C,
\label{eq:34}
\end{align}
where \(\mu _1(n)\) is the learning rates for controller 1 that adapts dynamically to the error, \(\mu _2(n)\) is the learning rate that is fixed for controller 2, \(a\) represents the sensitivity coefficient for error variation, \(\beta\) is the intensity coefficient for error change, \(\sigma \) is the regularization parameter for the learning rate of controller 1. 

\subsubsection{Controller weight vector update}
To enable the two controllers to adjust their weights based on error \(\textbf{e}(n)\) dynamically and enhance system stability, we introduce a weight update regularization parameter \(\varphi \) to prevent overfitting during the weight update process. We show the update of the controller weight below
\begin{align}
{\textbf{w}_1}(n + 1) = {\textbf{w}_1}(n) + \frac{{2{\mu _1}(n){\textbf{e}_1}(n)\textbf{x}(n)}}{{\varphi  + {\textbf{x}^T}(n)\textbf{x}(n)}},
\label{eq:35}
\end{align}
\begin{align}
{\textbf{w}_2}(n + 1) = {\textbf{w}_2}(n) + {\mu _2}{\textbf{e}_2}(n)\textbf{x}(n).
\label{eq:36}
\end{align}
where \({\textbf{w}_1}\) and \({\textbf{w}_2}\) denote the weight vectors for controller 1 and 2, respectively. 
They are dynamically updated to minimize the overall error \(\textbf{e} (n)\).

\subsubsection{Controller weight transfer}
\begin{figure}[htbp]
\centerline{\includegraphics[width=0.65\columnwidth]{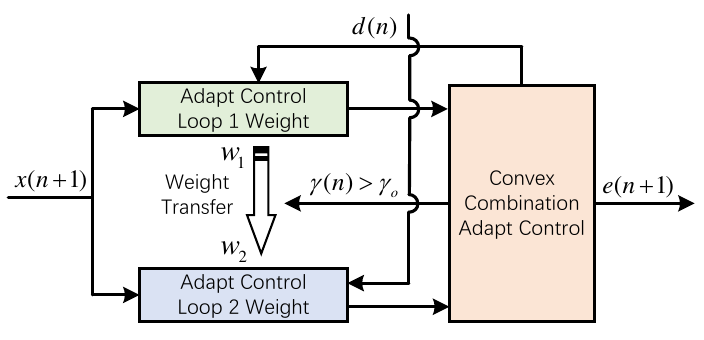}}
\caption{The schematic diagram for the weight transfer in the proposed coil control.}
\label{fig:5}
\end{figure}

The two controllers in~\Cref{fig:5} can compete with each other is they are not well coordinated and lead to reduced convergence of the control. To coordinate the controllers and improve the convergence of the entire coil control, we define a coupling coefficient \(\lambda(n)\) to facilitate the weight transfer between the controllers, as illustrated in \Cref{fig:5}. The weight transfer aims to adjust the two controllers to harmonize their performance as a whole toward the control objectives. We assume that the linear controller exhibits a faster convergence, and we define the weight transfer as
\begin{align}
{\textbf{w}_2}(n {+} 1) {=} \left\{\!\!\!\!\begin{array}{l}{\textbf{w}_1}(n + 1),~if~\gamma {{(n) > }}{\gamma _o}~and~n\bmod {T_o} {=} 0,\\{\textbf{w}_2}(n + 1),~otherwise,\end{array} \right.
\label{eq:37}
\end{align}
where \(T_o\) represents the weight update period specified according to the magnetometer sample rate. It holds that the longer the update period is, the slower the weight transfer will be. \({\gamma _o}\) is the weight transfer threshold and \(0 \ll {\gamma _o} < 1\). Note that when the coupling coefficient is larger than the threshold, the controller weight vectors will undergo a transfer.

\subsubsection{Update factor iteration}
We renew the update factor to reflect the interaction between the two controllers over time. We trigger the renew of the update factor \(b(n)\) when the weight transfer threshold \(\gamma {{(n) > }}{\gamma _o}\) is met. We conduct the update iteratively using the least mean squares method~\cite{DBLP:journals/tnn/ZhangCYZH18}, and the update factor is calculated as follows
\begin{align}
b(n {+} 1) {=} b(n) {+} {\mu _b}sign\{ \!\textbf{e}(n)\!\} ({\textbf{y}_1}(n) {-} {\textbf{y}_2}(n))\gamma (n)(1 {-} \gamma (n)),
\label{eq:38}
\end{align}
where \(\mu_b\) is the learning rate of the iteration coefficient, and \(sign\) is the sign function.

Note that by integrating the error that indicates how the generated field diverges from the target value, the designed coil control can compensate the current against disturbances from the surrounding environment.

With the six processes, the convex combination coil control can provide the current flowing through the coil that leads to stable and accurate magnetic field generation with sound convergence. We provide convergence and stability proofs in the following sections.


\subsection{Convergence proof}
\label{sec:3.2}
\textit{\textbf{Proof:}} Assume that both the desired signal \(\textbf{d}(n)\) and input signal \(\textbf{x}(n)\) are of stationary random processes. We also assume the optimal controller weight vector to be \(w_o\). The input signals \(\textbf{x}(n)\) and \(\textbf{x}(n+1)\) are hypothesized to be independent and uncorrelated. The weight error \(\textbf{v}(n)\), learning rate \(\mu(n)\), and weight vector \(\textbf{w}(n)\) are considered to be linearly independent and mutually independent. To ensure the convergence of the magnetic field generation, the maximum learning rate \(\mu\) of the controllers must satisfy the following condition~\cite{Flammini_2007}:
\begin{align}
0 < {\mu _{\max }} < \frac{2}{{{\lambda _{\max }}}},
\label{eq:39}
\end{align}
where \(\lambda _{\max}\) is the largest eigenvalue of an input autocorrelation matrix \({\textbf{R}_{xx}} = E[{\textbf{x}^T}(n)\textbf{x}(n)]\). 

Given that the learning rate of controller 2 is a constant while that of controller 1 is updated iteratively, the condition in~\Cref{eq:39} can be satisfied if the learning rate of controller 1 holds for~\Cref{eq:39}. Let \(\bar \mu (n) = {\mu _1}(n)/(\varphi  + {\textbf{x}^T}(n)\textbf{x}(n))\) and \(\textbf{v}(n) = \textbf{w}(n) - {\textbf{w}_o}\) denote the learning rate and weight error, respectively, where \(\textbf{w}_o\) represent the optimal weight. Then, substitute these variables into~\Cref{eq:35} and it yields
\begin{equation}
\begin{aligned}
\textbf{v}(\!n{+}1\!) &{=}\textbf{v}(n){+} 2\bar \mu (n)\textbf{e}(n)\textbf{x}(n) \\&{=} \textbf{v}(n) {+} 2\bar \mu (n)({\textbf{x}^T}(\!n\!){\textbf{w}_o} {+}\boldsymbol{\varepsilon} (n) {-} {\textbf{x}^T}(\!n\!)\textbf{w}(n))\textbf{x}(n) \\&{=} \textbf{v}(n) {-} 2\bar \mu (n){\textbf{x}^T}(\!n\!)\textbf{x}(\!n\!)(\textbf{w}(\!n\!) {-} {w_o}) {+} 2\bar \mu (n)\boldsymbol{\varepsilon} (n)\textbf{x}(n) \\&{=} \textbf{v}(n)(I {-} 2\bar \mu (n){\textbf{x}^T}(n)\textbf{x}(n)) {+} 2\bar \mu (n)\boldsymbol{\varepsilon} (n)\textbf{x}(n).
\end{aligned}
\label{eq:40}
\end{equation}

Taking the expectation of both sides of~\Cref{eq:40} and assuming that \(E[\boldsymbol{\varepsilon}(n)] = 0\) yields
\begin{align}
E\left[ {\textbf{v}(n + 1)} \right] = E\left[ {\textbf{v}(n)} \right](\textbf{I} - 2\bar \mu (n)E\left[ {{\textbf{x}^T}(n)\textbf{x}(n)} \right]).
\label{eq:41}
\end{align}

Since \({\textbf{x}^T}(n)\textbf{x}(n)\) is an autocorrelation matrix, there exists a unitary matrix \(\textbf{Q}\) such that the following equation holds
\begin{align}
{\textbf{R}_{xx}} = E[{\textbf{x}^T}(n)\textbf{x}(n)] = {\textbf{Q}^T}\Lambda \textbf{Q},
\label{eq:42}
\end{align}
where \(\boldsymbol{\Lambda} {=} diag[{\lambda _1}\!,\!{\lambda _2},\!{\ldots}\! {\lambda _N}\!]\), and \(\lambda\) denotes the eigenvalues of autocorrelation matrix \(\textbf{R}_{xx}\). 

Substitute \(\textbf{R}_{xx}\) into~\Cref{eq:41}, and we obtain
\begin{equation}
\begin{aligned}
E\left[ \textbf{v}(n + 1) \right] &= E\left[ \textbf{v}(n) \right](\textbf{Q}^T\textbf{Q} - 2\bar \mu (n)\textbf{Q}^T\boldsymbol{\Lambda} \textbf{Q}) \\
&= E\left[ \textbf{v}(n) \right](\textbf{Q}^T(\textbf{I} - 2\bar \mu (n)\boldsymbol{\Lambda})\textbf{Q}).
\end{aligned}
\label{eq:43}
\end{equation}

Applying the \(L2\) norm to~\Cref{eq:43} and using the Cauchy--Schwarz inequality~\cite{steele2004cauchy} yields
\begin{equation}
\begin{aligned}
\left\| E\left[ \textbf{v}(n {+} 1) \right] \right\|_2 &\le \left\| \textbf{Q}^T \right\|_2 \left\| \textbf{I} {-} 2\bar \mu (n)\boldsymbol{\Lambda} \right\|_2 \left\| \textbf{Q} \right\|_2 \left\| E\left[ \textbf{v}(n) \right] \right\|_2 \\
&= \left\| \textbf{I} - 2\bar \mu (n)\boldsymbol{\Lambda} \right\|_2 \left\| E\left[ \textbf{v}(n) \right] \right\|_2,
\end{aligned}
\label{eq:44}
\end{equation}

\begin{equation}
\textbf{I} {-} 2\bar \mu (\!n\!)\boldsymbol{\Lambda} {=}\! \begin{bmatrix}
1 {-} 2\bar \mu (\!n\!)\lambda_1 & 0 & \cdots & 0 \\ \!
0 & 1 {-} 2\bar \mu (\!n\!)\lambda_2 & \cdots & 0 \\ \!
\vdots & \vdots & \ddots & \vdots \! \\ \!
0 & 0 & \cdots & 1 {-} 2\bar \mu (\!n\!)\lambda_N
\end{bmatrix}\!\!,
\label{eq:45}
\end{equation}
where when \(\left| {1 - 2\bar \mu (k){\lambda _{\max }}} \right| < 1\), it holds that
\begin{align}
0 < \frac{\beta }{{\varphi  + {\textbf{x}^T}(n)\textbf{x}(n)}} \le \frac{2}{{{\lambda _{\max }}}}.
\label{eq:46}
\end{align}

Thus, the proposed method can ensure convergence of the field generation under the proposed coil control, by adjusting the values of error variation intensity coefficient \(\beta\) and weight update regularization parameter \(\varphi\). 

\subsection{Stability proof}
\label{sec:3.3}

In the learning rate update in~\Cref{eq:36}, the controller 1 is adapted according to the error value \(\textbf{e}_1 (n)\) , which includes noise \(\boldsymbol{\varepsilon} (n)\) and offset weight \(\Delta \textbf{w}(n) = \textbf{w}(n) - {\textbf{w}^*}\). Note that sudden changes in noise might induce unstable control by controller 1. Therefore, we need to ensure stability of the field generation when updating the learning rate.

\textit{\textbf{Proof:}}  From~\Cref{eq:31}, we can obtain
\begin{align}
\textbf{e}(n) = \boldsymbol{\varepsilon}(n) - {\textbf{x}^T}(n) \Delta \textbf{w} (n),
\label{eq:47}
\end{align}
\begin{align}
{\textbf{e}^2}\!(\!n\!) {=} {\boldsymbol{\varepsilon}^2}\!(n) {-} 2\boldsymbol{\varepsilon}(n){\textbf{x}^T}\!\!(n)\Delta \!\textbf{w} (n) {+} {\Delta \textbf{w}^T}\!\!(n)\textbf{x}(n){\textbf{x}^T}\!\!(n)\Delta \!\textbf{w} (n),
\label{eq:48}
\end{align}
\begin{align}
\textbf{e}(n - 1) =\boldsymbol{\varepsilon}(n - 1) - {\textbf{x}^T}(n - 1)\Delta \textbf{w}(n-1),
\label{eq:49}
\end{align}
thus, it holds that
\begin{equation}
\begin{aligned}
\textbf{e}(\!n\!)\textbf{e}(\!n{-}1\!)&= \boldsymbol{\varepsilon}(n)\boldsymbol{\varepsilon}(n {-}1){-}\boldsymbol{\varepsilon}(n){\textbf{x}^T}(n {-} 1)\Delta \textbf{w}(n{-} 1) \\
&\quad\!\!\!{-}\textbf{x}(\!n\!)\Delta \textbf{w} (\!n\!)\boldsymbol{\varepsilon}(\!n{-}1\!){+}{\textbf{x}\!^T}\!\!(\!n\!)\Delta \textbf{w}(\!n\!){\textbf{x}\!^T}\!\!(\!n{-}1\!)\Delta \textbf{w}(\!n{-}1\!)\!.
\end{aligned}
\label{eq:50}
\end{equation}

Since noise \(\boldsymbol{\varepsilon}(n)\) is a zero-mean Gaussian noise and is independent of input \(\textbf{x}(n)\), taking the expectation of~\Cref{eq:48}~and~\Cref{eq:50} yields:
\begin{align}
E[{\textbf{e}^2}(n)] = E[{\boldsymbol{\varepsilon}^2}(n)] + E[{\Delta \textbf{w}^T}(n)\textbf{x}(n){\textbf{x}^T}(n)\Delta \textbf{w}(n)],
\label{eq:51}
\end{align}
\begin{align}
E[\textbf{e}(n)\textbf{e}(n - 1)] = E[\Delta \textbf{w}(n)\textbf{x}(n){\textbf{x}^T}(n - 1)\Delta \textbf{w}(n - 1)].
\label{eq:52}
\end{align}

According to~\Cref{eq:51}~and~\Cref{eq:52}, updating the learning rate of controller 1 based on the value of error \(\textbf{e}(n)\textbf{e}(n-1)\) can reduce the influence of noise \(\boldsymbol{\varepsilon}(n)\) on the control over time and lead to stability of the controllers, resulting in a stable field generation. \hfill $\blacksquare$

\section{Testbed implementation and validation}
\label{sec:Hils Performance Validation Methodology}
This section implements and validates the Hardware-in-the-loop simulation (Hils) testbed following the designed coil parameter optimization and convex combination coil control. We build the Hils testbed for geographic navigation using off-the-shelf hardware, and we test the testbed and analyze the quality of the magnetic field generation regarding uniformity, accuracy, stability, convergence, and rapid response the field generation. we also evaluate the coil control of the testbed in comparison with existing coil control methods.

\begin{figure}[htbp]
\centering
\subfigure[]
{
    \begin{minipage}[b]{0.45\linewidth}
        \centering
        \includegraphics[width=\textwidth, angle=-90]{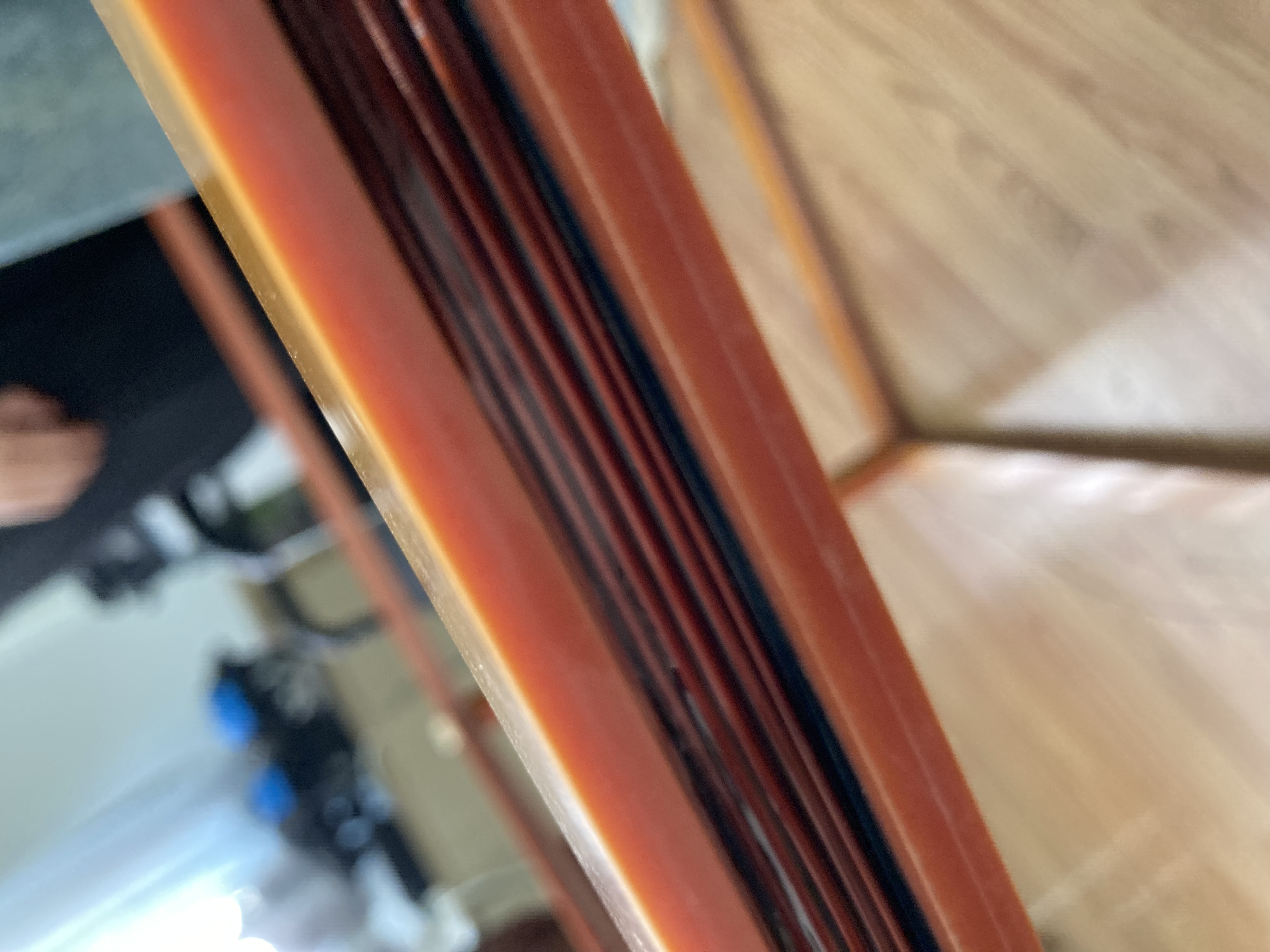}
        \vspace{3pt}
    \end{minipage}\label{fig:6a}
}
\subfigure[]
{
 	\begin{minipage}[b]{0.45\linewidth}
        \centering
        \includegraphics[width=\textwidth, angle=-90]{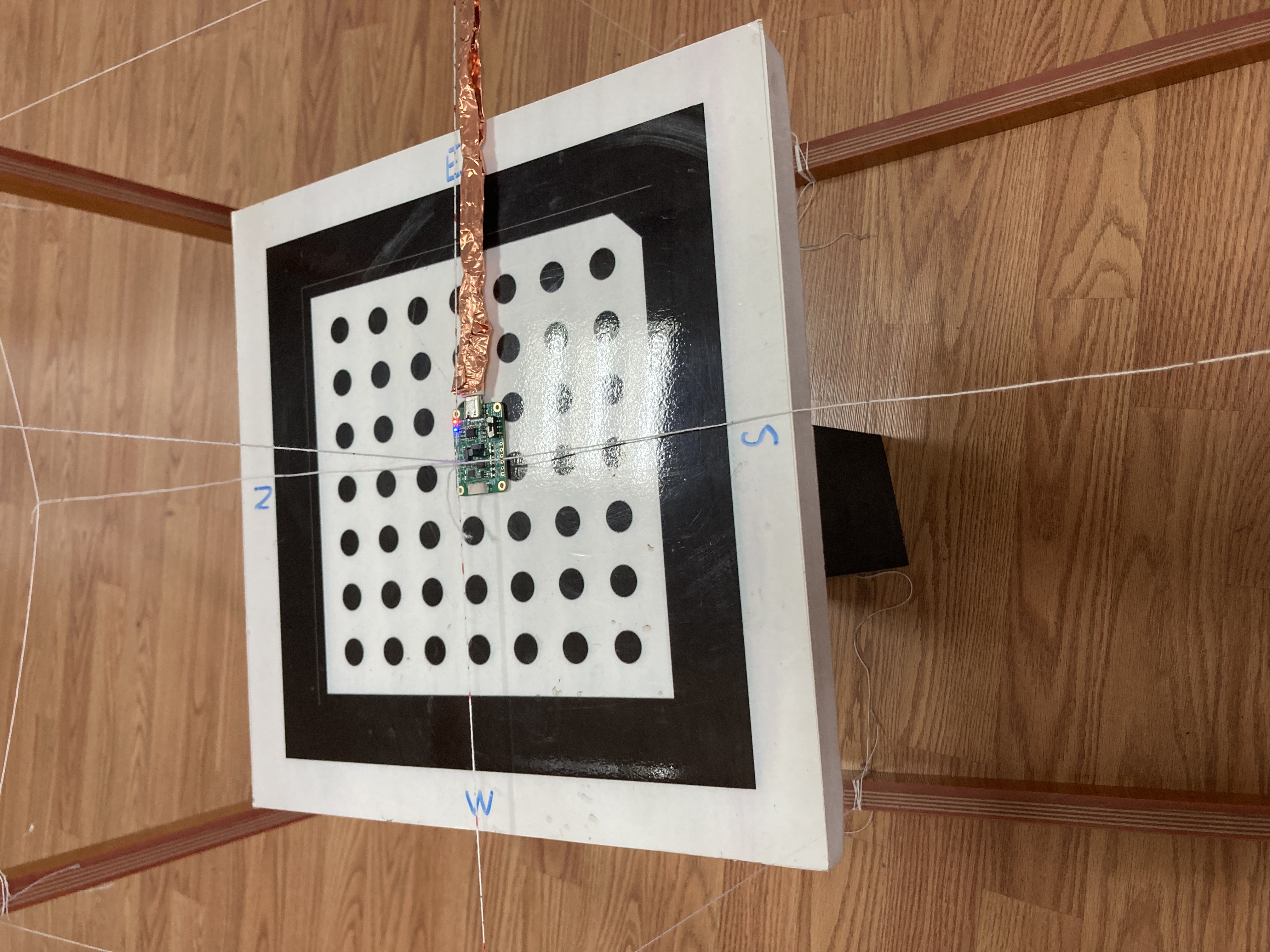}
        \vspace{3pt}
    \end{minipage}\label{fig:6b}
}
\caption{(a) The Helmholtz coil built in this study (b) Measurement diagram that records the field strength by the Helmholtz coil.}
\label{fig:6}
\end{figure} 

\subsection{Testbed hardware settings}

Following the theoretical design presented in \Cref{sec:Hardware Design and Implementation}, we construct the Helmholtz coil of the Hils testbed as shown in~\Cref{fig:6a}. The coil supports are crafted from a \(20~\text{mm}\)-thick bakelite, where each side of the coil is \(840.4~\text{mm}\), and the spacing between the two coils is \(457.6~\text{mm}\). The coil support has grooves on the outer side, which are \(6~\text{mm}\) in depth and \(9.5~\text{mm}\) in width. The winding wire is made of QZ/155 grade modified polyester enameled copper wire. To prevent the coil from overheating after long periods, we use a copper wire with a diameter of \(1.2~\text{mm}\), which can handle the peak current of up to \(5~\text{A}\). 
The coils are wound densely in a total of four layers, with six turns per layer, and are also secured and sealed with insulating tape. The final coil resistance is \(3.11~\Omega\), and the inductance is \(5.20~\mu \text{H}\).

Following the testbed system architecture in~\Cref{fig:2}, in the constructed testbed, the control computer (Dell Inspiron 14 Plus) sends the target commands to the microcontroller via RS232. The microcontroller (model STM32F4-Discovery) receives the magnetic field data returned by the feedback magnetic sensor (PNI RM3100). Then the control signal for the coil is calculated by measuring the difference between the current and target magnetic field strengths and implementing the convex combination control. The voltage-magnetic fitting curve is used to compute the control signal for the input digital-to-analog converter (analog devices AD5780). The voltage-controlled current source (Texas Instruments OPA549) is used to generate the current fed to the coils. This current then generates the uniform and stable magnetic field required by the geomagnetic navigation. The entire system was powered by two 12-V power sources (Mean Well LRS-100-12). We provide the specifications of all the components in the Hils testbed in details in~\Cref{tab:1}.

\begin{table}[ht]
\caption{Specifications of the testbed components}
\label{tab:1}
\setlength{\tabcolsep}{3pt}  
\renewcommand{\arraystretch}{1.1}  
\centering
\begin{tabular}{@{}lll@{}}  
\toprule
\textbf{Subsystems} & \textbf{Model} & \textbf{Parameters} \\ 
\midrule
\begin{tabular}[c]{@{}l@{}}Voltage-controlled \\ current source\end{tabular} &
  \begin{tabular}[c]{@{}l@{}}Texas Instruments\\ OPA549 (1 Pc.)\end{tabular} &
  \begin{tabular}[c]{@{}l@{}}Peak Output current: 10~A\\ Transconductance: 1~S\\ Power supply: ±4~V to ±30~V\end{tabular} \\ 
\midrule
Microcontroller &
  \begin{tabular}[c]{@{}l@{}}STM32F4-\\ Discovery (1 Pc.)\end{tabular} &
  \begin{tabular}[c]{@{}l@{}}Model type: ARM Cortex M4\\ Clock Freq: 180~MHZ\end{tabular} \\ 
\midrule
Power supply &
  \begin{tabular}[c]{@{}l@{}}Mean Well \\ LRS-100-12 (2 Pc.)\end{tabular} &
  \begin{tabular}[c]{@{}l@{}}Output voltage: 12~V\\ Output power: 100~W\\ Output current: 0--8.5~A\end{tabular} \\ 
\midrule
\begin{tabular}[c]{@{}l@{}}Digital-to-\\ analog converter\end{tabular} &
  \begin{tabular}[c]{@{}l@{}}Analog devices \\ AD5780 (1 Pc.)\end{tabular} &
  \begin{tabular}[c]{@{}l@{}}Resolution: 18~Bit\\ Power supply: ±3.5~V to ±16.5~V\\ Differential Nonlinearity: ±1~LSB\\ Reference voltage: ±5~V\end{tabular} \\ 
\midrule
\multirow{5}{*}{Magnetic sensor} &
  \begin{tabular}[c]{@{}l@{}}PNI \\ RM3100 (2 Pc.)\end{tabular} &
  \begin{tabular}[c]{@{}l@{}}Sensitivity: 13~\(n\)T\\ Measurement range: ±800~uT\\ Noise: 15~\(n\)T\\ Max sample rate: 200~Hz\end{tabular} \\ 
 & \begin{tabular}[c]{@{}l@{}}Pixhawk 2.4.6\\ HMC5883L (2 Pc.)\end{tabular} &
  \begin{tabular}[c]{@{}l@{}}Sensitivity: 435~nT/LSB\\ Measurement range: ±800~uT\\ Noise: 200~\(n\)T\\ Max sample rate: 75~Hz\end{tabular} \\ 
\midrule
Control computer &
  \begin{tabular}[c]{@{}l@{}}Dell Inspiron \\ 14 Plus (1 Pc.)\end{tabular} &
  \begin{tabular}[c]{@{}l@{}}CPU: I5-13420H\\ Graphic: RTX 3050\end{tabular} \\ 
\bottomrule
\end{tabular}
\end{table}

\subsection{Validation of magnetic field uniformity}
\label{sec:4.1}
This section evaluates the uniformity of the magnetic field generated near the center of the Helmholtz coil in the designed testbed. We also examine how the generated field by the coil is aligned with its virtual counterpart, simulated by Finite Element Method (FEM). Particularly, we use the Ansys Maxwell Electronics FEM to simulate a uniaxial square Helmholtz coil in three-dimensional model. We employ the magnetostatic solver to simulate the magnetic field distribution in FEM. We list the configuration discrepancies between the virtual and physical coil in~\Cref{tab:2}, which is inevitable due to measurement errors and a limited choice of real materials in constructing the Helmholtz coil.  
The results show discrepancies of field uniformity between the field generated by Helmholtz coil and its virtual counterpart by FEM, but such discrepancies is confined in a narrow range. Below we elaborate the field uniformity validation and result analysis. 

\begin{table}[ht]
\centering
\caption{The parameters of the virtual coil by FEM and those of the physical coil.}
\label{tab:2}
\begin{tabularx}{0.7\linewidth}{lcc}
\toprule
\textbf{Parameter} & \textbf{FEM Settings} & \textbf{Actual Hils} \\ 
\midrule
Coil length \(L\)~(mm) & 840.4 & 846.0 \\ 
Coil spacing \(d\)~(mm) & 457.6 & 458.0 \\ 
Current \(I\)~(A) & 2.94 & 2.93 \\ 
Coil turns \(N\)~(Turn) & 24 & 24 \\ 
Material & Copper & QZ/155 \\ 
Series inductance (\(\mu\)\text{H}) & 4.72 & 5.20 \\ 
Calculate region (\%) & 200 & - \\ 
Maximum number of passes & 30 & - \\ 
Percent error (\%) & 0.1 & - \\ 
Refinement per pass (\%) & 30 & - \\ 
\bottomrule
\end{tabularx}
\end{table}

\begin{figure}[htbp]
\centering
\subfigure[]
{
    \begin{minipage}[b]{0.33\linewidth}
        \centering
        \includegraphics[width=\textwidth]{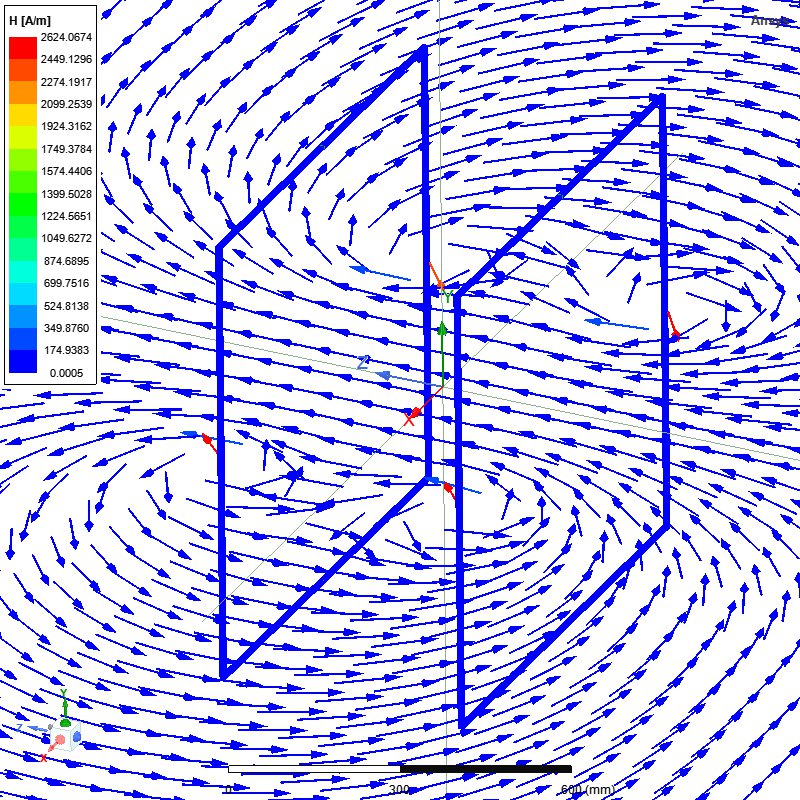}
    \end{minipage}\label{fig:7a}
}
\subfigure[]
{
 	\begin{minipage}[b]{0.33\linewidth}
        \centering
        \includegraphics[width=\textwidth]{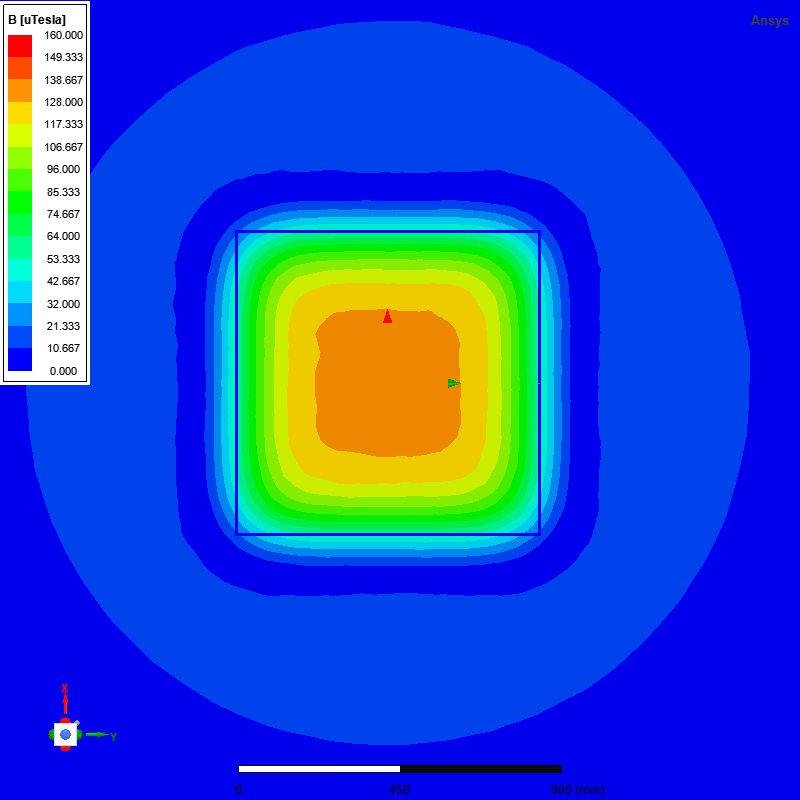}
    \end{minipage}\label{fig:7b}
}
\subfigure[]
{
 	\begin{minipage}[b]{0.33\linewidth}
        \centering
        \includegraphics[width=\textwidth]{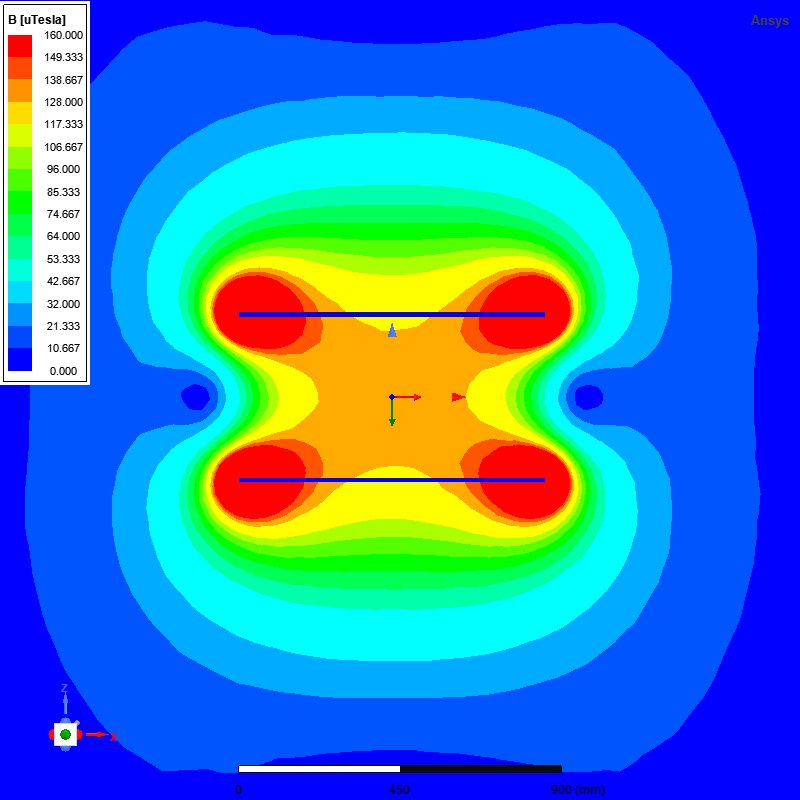}
    \end{minipage}\label{fig:7c}
}
\subfigure[]
{
 	\begin{minipage}[b]{0.33\linewidth}
        \centering
        \includegraphics[width=\textwidth]{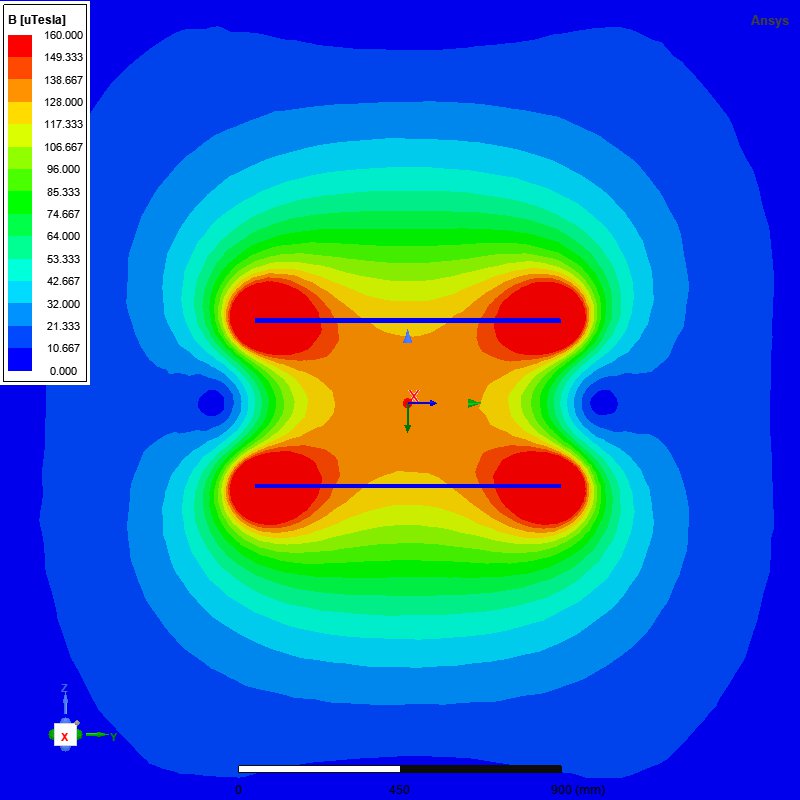}
    \end{minipage}\label{fig:7d}
}
\caption{Field strength of the virtual coil simulated by FEM. (a) The simulated coil with its magnetic field near the coil, (b) Field strength on the XOY plane, (c) Field strength on the XOZ plane, (d) Field strength on the YOZ plane.}
\label{fig:7}
\end{figure}

We present the virtual field generate by FEM in \Cref{fig:7}, where we observe that the magnetic field lines are parallel near the center of the coil. The magnetic field strength across sections indicated that a uniform magnetic field is generated across the XOY, XOZ, and YOZ planes, as shown in~\Cref{fig:7b},~\Cref{fig:7c}, and \Cref{fig:7d}. The field strength in all three planes is orthogonal and no coupling is observed. 

\begin{figure}[htbp]
\centerline{\includegraphics[width=0.55\columnwidth]{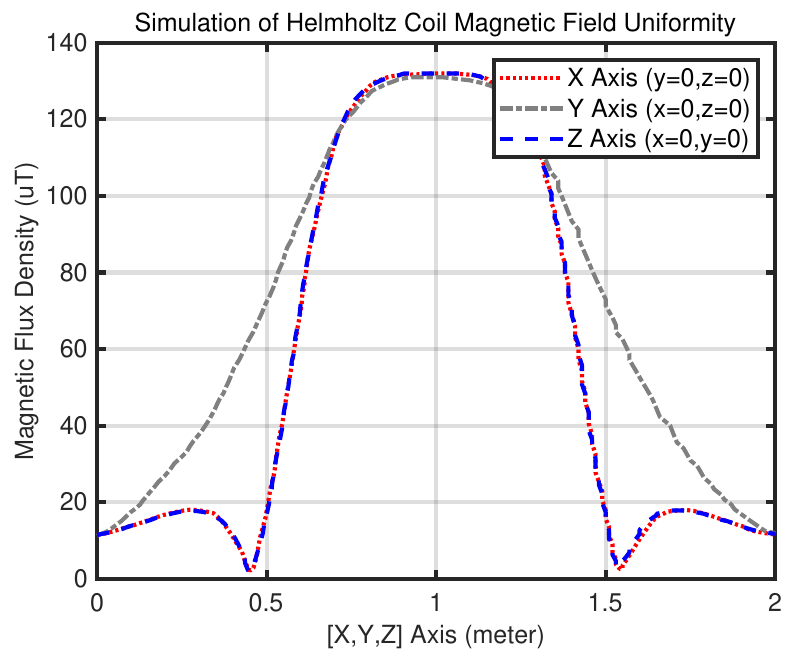}}
\caption{The magnetic flux density distribution across x-, y-, and z-axis directions.}
\label{fig:8}
\end{figure}

We also measure the field strength distribution across the three directional planes, as shown in \Cref{fig:8}. We observe the magnetic field distribution in the \textit{Y}-axis direction is slightly different from the \textit{X}-axis and \textit{Z}-axis. Around the coil center spanning \(0.75~\text{m}-1.25~\text{m}\), the field strength is greater than \(120~\mu \text{T}\) and is relatively uniform in its distribution, which meets the design requirements for the size of the carrier (a Pixhawk 2.4.6 controller with the size \(80~\text{mm} \times 50~\text{mm} \times 15~\text{mm}\)in this study). Below we analyze the field generation of the physical Helmholtz coil and check its alignment with the virtual field by FEM.

\begin{figure}[htbp]
\centerline{\includegraphics[width=0.75\columnwidth]{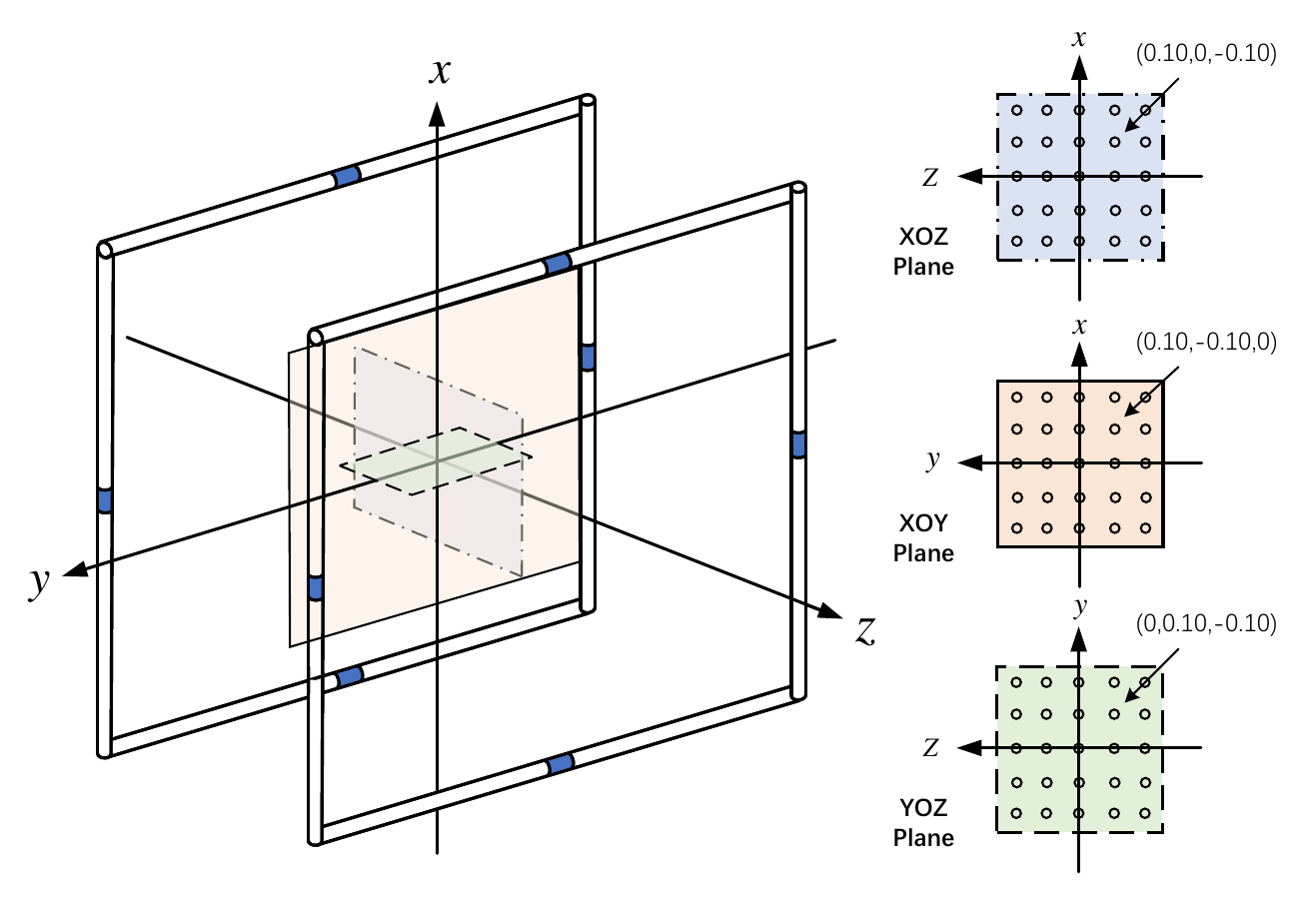}}
\caption{The field strength measurement schematic diagram for the physical coil.}
\label{fig:9}
\end{figure}

We place the physical Helmholtz coil at the geographic local at \((108.946466~E^\circ , 34.347269~N^\circ )\). We define the magnetic field in the northward direction as positive along the \textit{x}-axis, and we define the eastward direction as positive along the \textit{y}-axis, and the vertically upward direction as positive along the \textit{z}-axis. We measure the actual field strength at the determined location, and the intensity of the \(B_x\) component is 29,950.1~\(n\text{T}\), the \(B_y\) component is 21,290.2~\(n\text{T}\), and the \(B_z\) component is -51,917.4~\(n\text{T}\). We place sensors on three planes, and we record the measured field strength data at 25 points on each plane. We take 200 data points per point and take their average to reduce anomalies. The Helmholtz coil is oriented eastward during the measurement (along the \textit{y}-axis). We put and fix the magnetic sensors at the center of the coil. We conduct the field measurement using the PNI RM3100 tri-axial magnetic sensor, employing the data recording approach shown in \Cref{fig:9}.

\begin{figure}[htbp]
\centering
\subfigure[]
{
    \begin{minipage}[b]{0.55\linewidth}
        \centering
        \includegraphics[width=\textwidth]{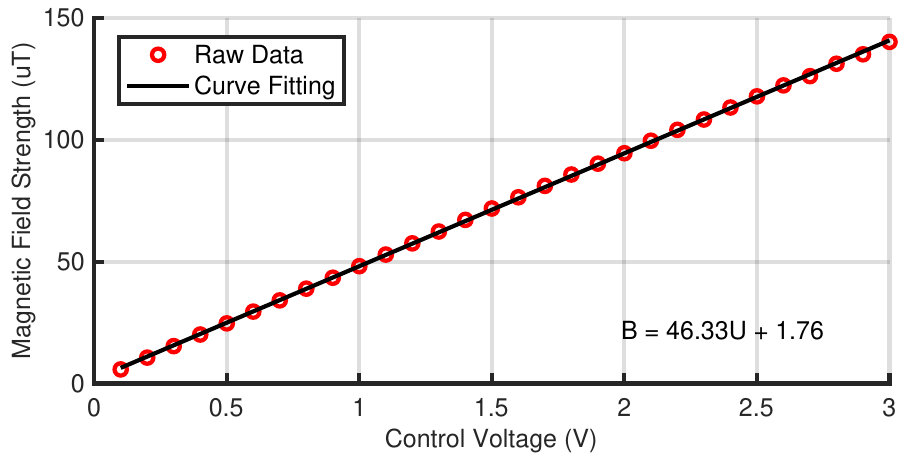}
    \end{minipage}\label{fig:10a}
}
\subfigure[]
{
 	\begin{minipage}[b]{0.55\linewidth}
        \centering
        \includegraphics[width=\textwidth]{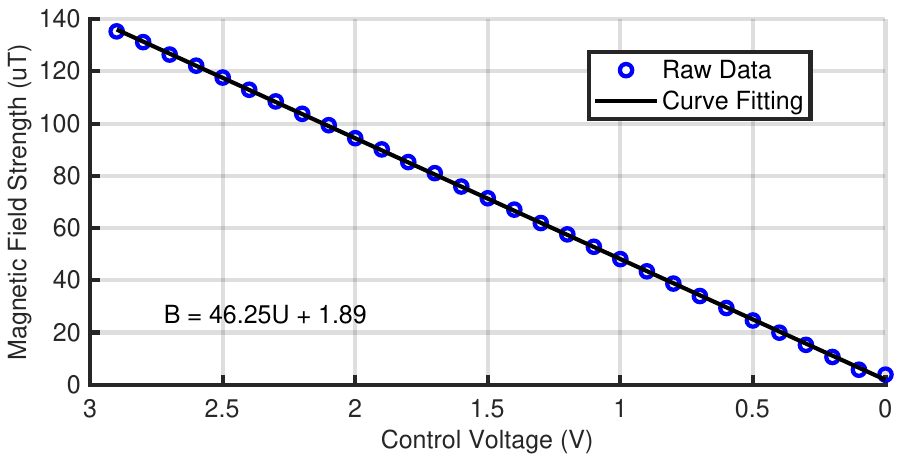}
    \end{minipage}\label{fig:10b}
}
\caption{Fitted curves for the mapping between voltage and magnetic field strength. (a) Fitted curve with ascending trend, (b) fitted curve with descending trend.}
\label{fig:10}
\end{figure}

This study used the digital-to-analog converter module to output driving voltages within the range from zero to 3~V. The driving voltage determines the current fed to the coil. 
The magnetic field generated by the coil is continuously recorded for 1~s, and the average value of 200 data points during the recording is taken as the final result. The \(B_y\) component of the magnetic sensor is recorded when the current flowed through the coil. Since the strength of the magnetic field generated by the coil is directly proportional to the driving voltage, a linear relationship between the control voltage and the target magnetic field is determined through linear fitting 
, as shown in \Cref{fig:10}. 

\begin{figure}[htbp]
\centering
\subfigure[]
{
    \begin{minipage}[b]{0.4\linewidth}
        \centering
        \includegraphics[width=\textwidth]{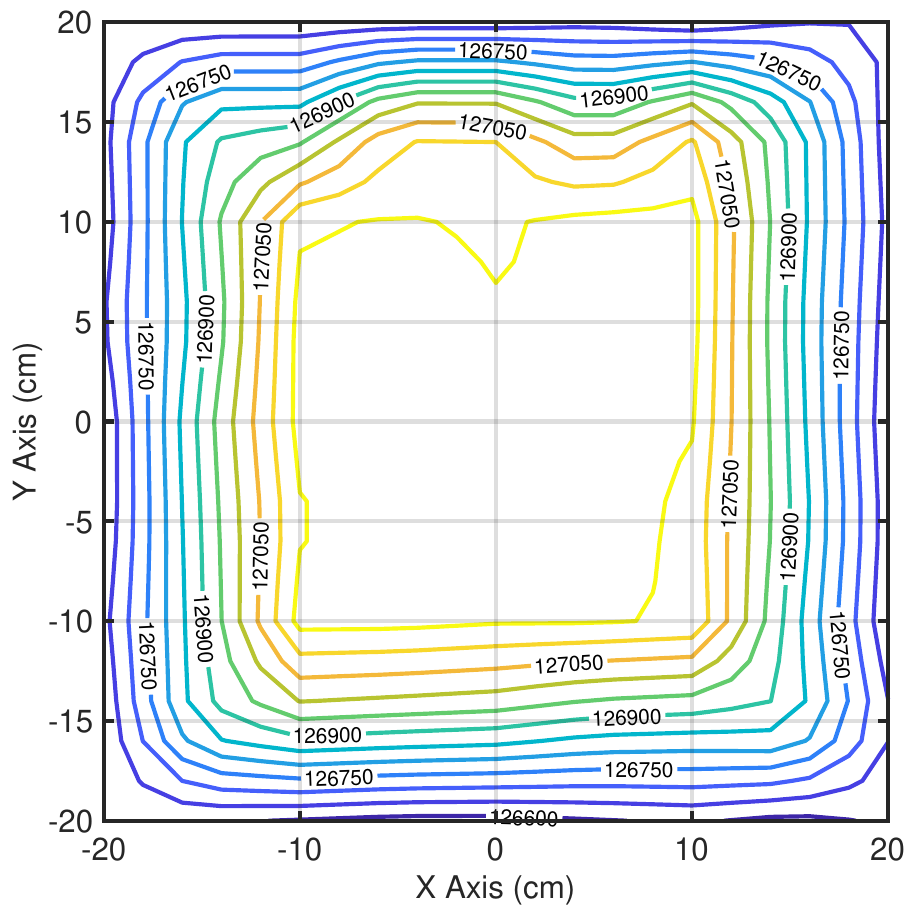}
    \end{minipage}
    \label{fig:11a}
}
\subfigure[]
{
 	\begin{minipage}[b]{0.4\linewidth}
        \centering
        \includegraphics[width=\textwidth]{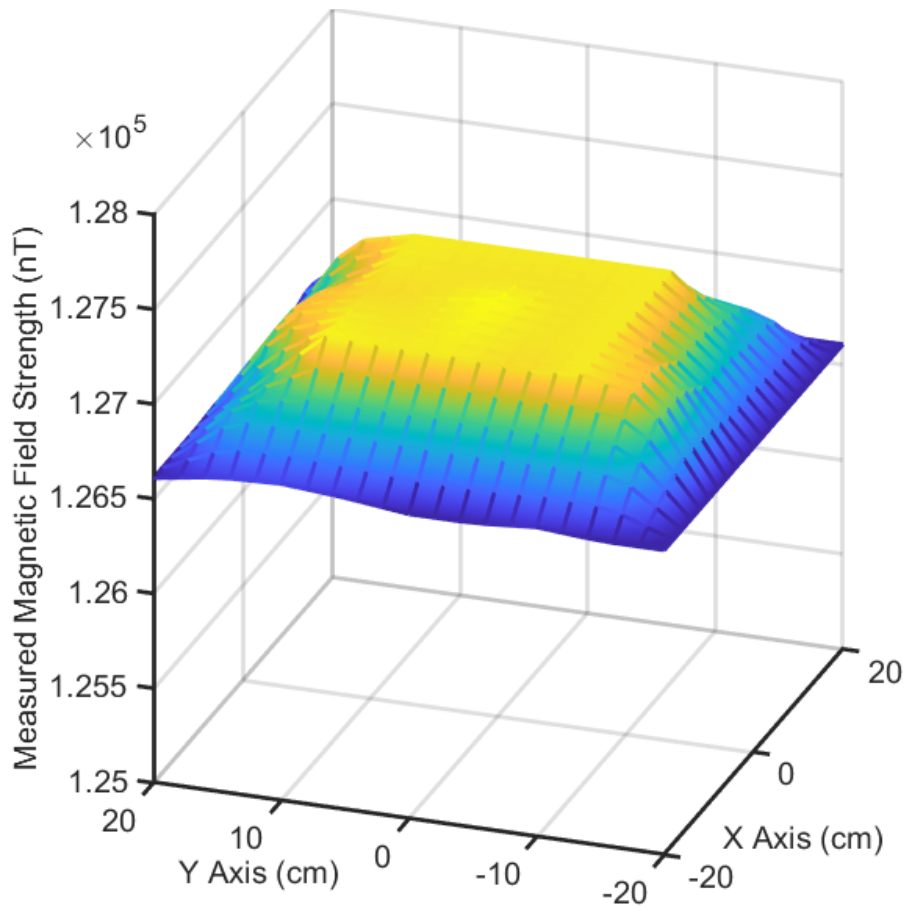}
    \end{minipage}
    \label{fig:11b}
}
\caption{(a) Field strength contour on XOY plane, (b) Field strength distribution in a three-dimensional space.}
\label{fig:11}
\end{figure}

Following our designed coil control, the values of parameters \(k\) and \(b\) were obtained using \Cref{eq:26}. The fitting parameters are determined as \(k=46.333\) and \(b=1.7623\) for \Cref{fig:10a} where the voltage increase, and \(k=46.253\) and \(b=1.8935\) for \Cref{fig:10b} where the voltage decreases. Based on the measurement and fitting results, the uniformity of the magnetic field generated by the Helmholtz coil is analyzed. We obtain the field uniformity on Plane B by \Cref{eq:17} and display it in \Cref{tab:3}. In the table, \(x\) represents the distance from the central origin to the test point on the \textit{x}-axis, \(y\) is the distance from the central origin to the test point on the \textit{y}-axis, and \(d\) is the space between the two coils. The magnetic field strength contour intensity on the XOY plane obtained from actual measurements is displayed in \Cref{fig:11}.

The results in \Cref{tab:3} indicates that near the center of the Helmholtz coil in our Hils testbed, the area with 5\% uniformity of the field strength can cover an area of approximately \(467~\text{mm} \times 467~\text{mm}\), which is sufficient for a carrier (in our design, we consider an autonomous underwater vehicle of BlueROV with the size of \(457~\text{mm} \times 338~\text{mm} \times 254~\text{mm}\)). Though we observe the uniformity variations along the \textit{x}-axis and \textit{y}-axis, those variations exist only at the edge of the uniformity area and they are considered to have a minimal impact on the experimentation. These discrepancies are attributed to the deviations during the assembly and manufacturing processes. The FEM and physical coil measurements align well with each other, with a maximum magnetic field error of 3\%, indicating that the actual manufactured coil works closely to its virtual counterpart. Furthermore, the field uniformity of the our testbed within an \(80~\text{mm} \times 80~\text{mm}\) around the center is less than \(H \le 0.5\% \), which is sufficient to accommodate the the controller (in our design, we use Pixhawk 2.4.6 controller with the size of \(80~\text{mm} \times 50~\text{mm} \times 15~\text{mm}\)). 

\begin{table}[ht]
\centering
\caption{Summary of the Helmholtz coil magnetic field uniformity.}
\label{tab:3}
\begin{tabularx}{0.75\linewidth}{l *{6}{c}}
\toprule
\diagbox[width=8em]{\textbf{D}}{\textbf{H}} & \textbf{0.1\%} & \textbf{0.5\%} & \textbf{1\%} & \textbf{5\%} & \textbf{10\%} & \textbf{20\%} \\ 
\midrule
\( \pm x/d\) & 0.059 & 0.125 & 0.319 & 0.515 & 0.602 & 0.720 \\ 
\( \pm y/d\) & 0.058 & 0.122 & 0.316 & 0.513 & 0.599 & 0.715 \\ 
\bottomrule
\end{tabularx}
\end{table}

\subsection{Field generation validation in a shielded environment}

This section conduct experiments with our testbed and examines the field generation in a shielded lab. We check how the testbed works without external inference, and we look into the field generation regarding the quality indicators of accuracy, stability, and convergence, and rapid response under the proposed convex combination coil control. We also compare our coil control with the state-of-the-art coil control approaches, including the LMS~\cite{Flammini_2007}, SVS \cite{Bendoumia_2024} and ATLMS~\cite{Jia_2023} methods. Amongst those methods, LMS serves as the baseline for assessing convergence and stability; SVS is characterized by its variable step-size adaptation that can promote robustness and adaptability against dynamic noise conditions in the surrounding environment; ATLMS utilizes an arctangent function for step-size modulation that can mitigate the impact of noise in the surrounding environment.


\begin{figure}[htbp]
\centering
\subfigure[]
{
    \begin{minipage}[b]{0.45\linewidth}
        \centering
        \includegraphics[width=\textwidth]{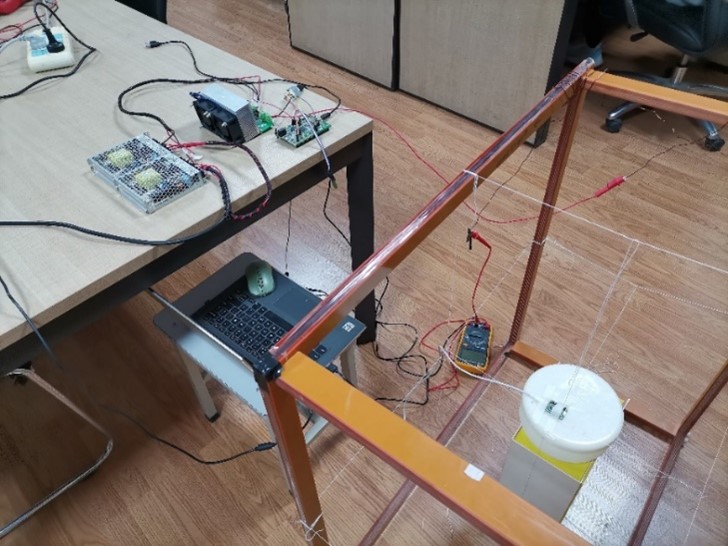}
    \end{minipage}
    \label{fig:14a}
}
\subfigure[]
{
 	\begin{minipage}[b]{0.45\linewidth}
        \centering
        \includegraphics[width=\textwidth]{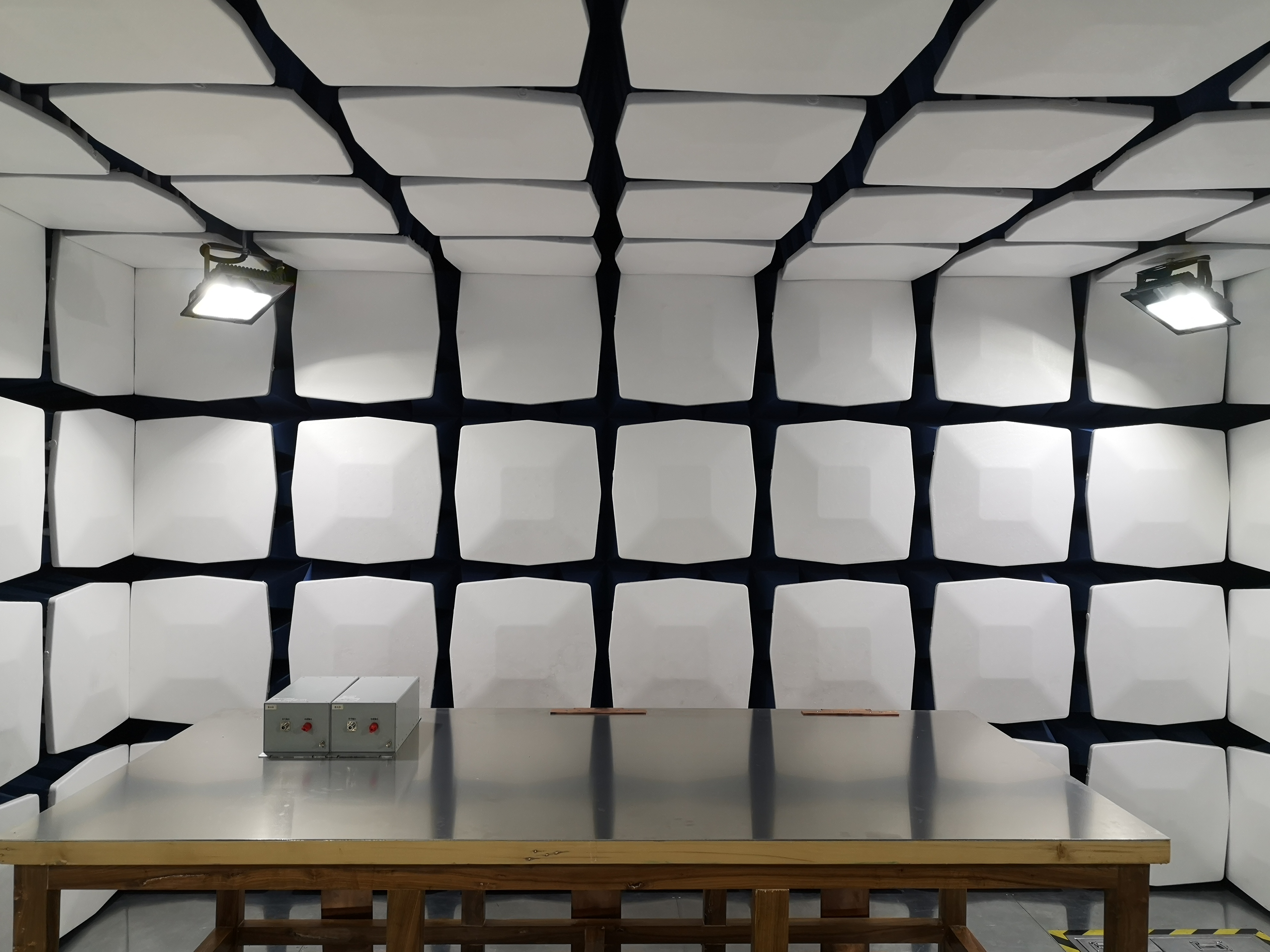}
    \end{minipage}
    \label{fig:14b}
}
\caption{Experiments of field generation in different environments: (a) An unshielded laboratory, (b) a shielded environment.}
\label{fig:14}
\end{figure}

\begin{figure}[htbp]
\centering
\subfigure[]
{
    \begin{minipage}[b]{0.45\linewidth}
        \centering
        \includegraphics[width=\textwidth]{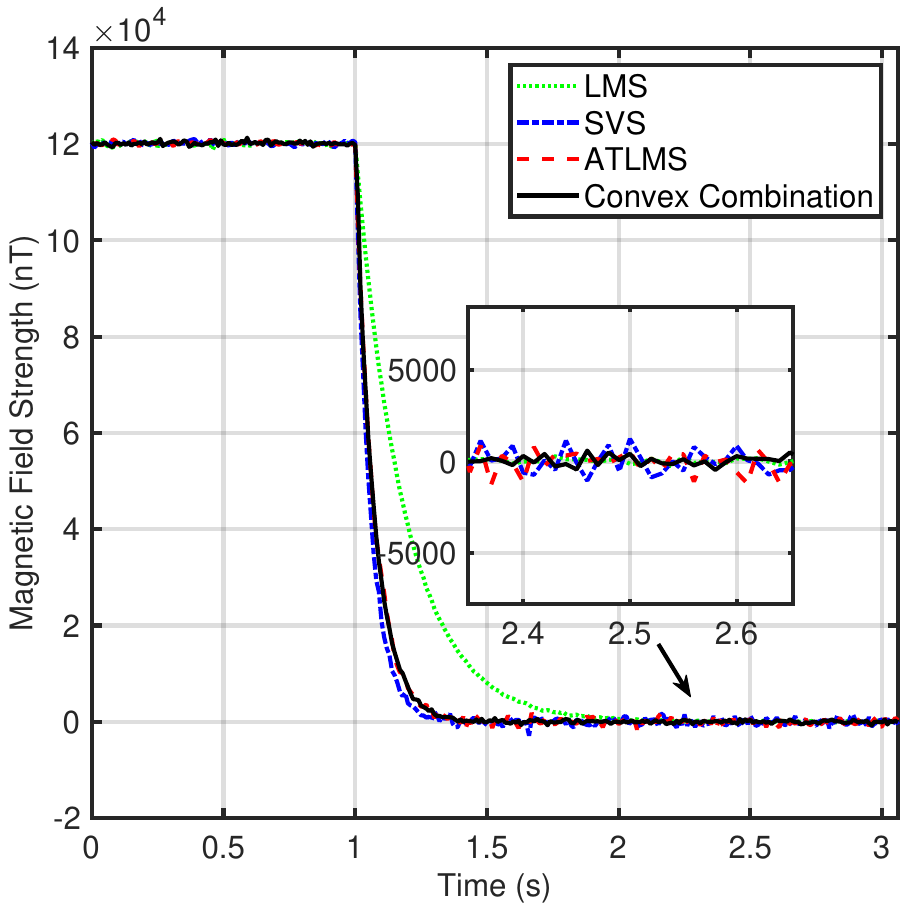}
    \end{minipage}
    \label{fig:15a}
}
\subfigure[]
{
 	\begin{minipage}[b]{0.45\linewidth}
        \centering
        \includegraphics[width=\textwidth]{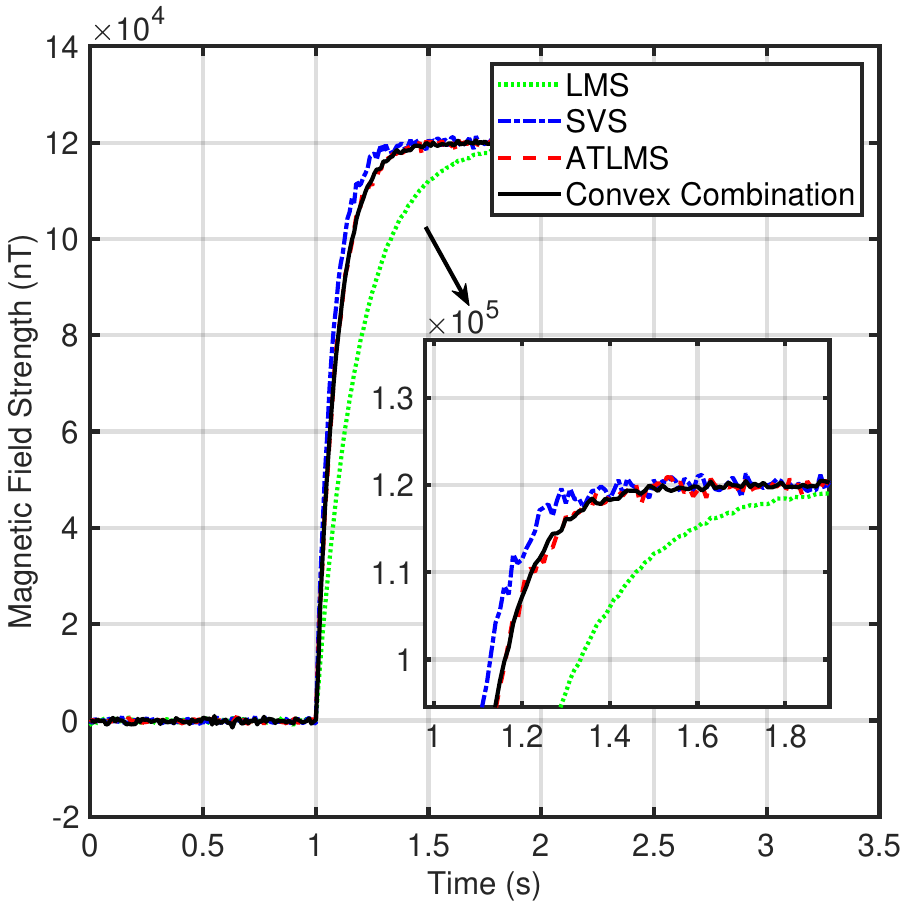}
    \end{minipage}
    \label{fig:15b}
}
\caption{The magnetic field control results. (a) Controlled field strength with the target from \(120,000~n\text{T}\) to zero, (b) controlled field strength with the target from zero to \(120,000~n\text{T}\).}
\label{fig:15}
\end{figure}

We carry out this experiment in an electromagnetic shielding room, as shown in \Cref{fig:14b}. The data collection process was performed using the HMC5883L embedded in Pixhawk 2.4.6., which is equipped with an autonomous underwater vehicle (BlueROV). We set the target field as (i) from \(120,000~n\text{T}\) to zero shown in \Cref{fig:15a}, and (ii) an ascending curve from zero to \(120,000~n\text{T}\) shown in \Cref{fig:15b}. We set the dynamic target field to test how the generated field can rapidly respond and follow the target field. We set the parameters of the control approaches to their optimal values according to the conditions of a high SNR of \(30~\text{dB}\), as displayed in \Cref{tab:4-0}. We visualize the field generation in \Cref{fig:14} and we provide the numerical results in \Cref{tab:7}.

\begin{table}[ht]
\centering
\caption{The experiment parameter settings}
\label{tab:4-0}
\begin{tabularx}{0.65\linewidth}{c c X}
\toprule
\textbf{SNR} & \textbf{Method} & \textbf{Parameters} \\ 
\midrule
\multirow{4}{*}{\(30~\text{dB}\)} & LMS & \(\mu {=} 0.005\) \\ 
 &  SVS & \(\alpha {=} 4\), \(\beta {=} 0.15\) \\ 
 &  ATLMS & \(\alpha {=} 500\), \(\beta {=} 0.01\), \(m {=} 900\), \(n {=} 500\) \\ 
 &  Convex & \(\alpha {=} 500\), \(\beta {=} 0.01\), \(\gamma(0) {=} 0.5\), \(\gamma_o {=} 0.55\) \\ 
\bottomrule
\end{tabularx}
\end{table}

We observe from the results in \Cref{fig:15a} that, the proposed coil control can track the shifted target field in \(0.43~\text{s}\). After the field strength stabilized at \(1.5~\text{s}\), the proposed control obtains an average magnetic field strength of \(-16.12~n\text{T}\), with a Root Mean Squared Error (RMSE) of 269.36 and fluctuations between \(-780~n\text{T}\) and \(+598~n\text{T}\). We also observe LMS achieves a higher accuracy with lower residual average of \(22.81~n\text{T}\) and lower RMSE of 138.23 with less fluctuations, however, LMS is slow in convergence and it requires \(1.12~\text{s}\) to reach the target. SVS is the fastest in convergence that tracks the shifted target field in \(0.35~\text{s}\), but it is with the average residual field strength of \(-18.91~n\text{T}\) and the highest RMSE of 696.08, with significant fluctuations from \(-2,002~n\text{T}\) to \(+1,976~n\text{T}\), indicating low stability and accuracy in field generation. ATLMS follows the target field in \(0.42~\text{s}\) with a residual average of \(25.57~n\text{T}\) and a high RMSE of 665.41. Its magnetic field fluctuates between \(-1,911~n\text{T}\) and \(+1,612~n\text{T}\), indicating low field generation accuracy and stability.

\begin{table}[htpb]
\setlength\tabcolsep{2pt}
\centering
\caption{Numerical results of the coil control in a shielded environment.}
\label{tab:7}
\begin{tabularx}{0.85\linewidth}{lccccX}
\toprule
\textbf{\begin{tabular}[c]{@{}c@{}}Hils\\ control \\ target\end{tabular}} & \textbf{Method} & \textbf{\begin{tabular}[c]{@{}c@{}}Reach \\ target \\ time (s)\end{tabular}} & \textbf{\begin{tabular}[c]{@{}c@{}}Mean magnetic \\ field \\ strength \((n\text{T})\)\end{tabular}} & \textbf{RMSE} & \textbf{\begin{tabular}[c]{@{}c@{}}Steady-state \\ fluctuation \\  range \((n\text{T})\)\end{tabular}} \\ 
\midrule
\multirow{4}{*}{\begin{tabular}[c]{@{}c@{}}120~\(u\text{T}\) to\\zero\end{tabular}} & LMS & 1.12 & 22.81 & 138.23 & -338 to +390 \\ 
 & SVS & 0.35 & -18.91 & 696.08 & -2,002 to +1,976 \\ 
 & ATLMS & 0.42 & 25.57 & 665.41 & -1,911 to +1,612 \\ 
 & Convex & 0.43 & -16.12 & 269.36 & -780 to +598 \\ 
\midrule
\multirow{4}{*}{\begin{tabular}[c]{@{}c@{}}zero to \\120~\(u\text{T}\)\end{tabular}} & LMS & 1.32 & 119,999.75 & 141.54 & 119,665 to 120,289 \\ 
 & SVS & 0.48 & 119,971.54 & 698.71 & 118,248 to 122,447 \\ 
 & ATLMS & 0.52 & 120,012.88 & 671.05 & 117,923 to 121,628 \\ 
 & Convex & 0.55 & 119,991.95 & 272.66 & 119,275 to 120,601 \\ 
\bottomrule
\end{tabularx}
\end{table}

We observe from the results in \Cref{fig:15b} that, the proposed control follows the change in field strength within only \(0.55~\text{s}\). After the field generation stabilized, the proposed control achieves an average field strength of \(119,991.95~n\text{T}\) with an RMSE of 272.66, with fluctuations between \(119,275~n\text{T}\) and \(120,601~n\text{T}\). LMS provides an average field strength of \(119,999.75n\text{T}\) and an RMSE of 141.54 with fluctuations ranging from \(119,665~n\text{T}\) to \(120,289~n\text{T}\), indicating higher accuracy and stability in the field generation. However, LMS takes \(1.32~\text{s}\) to converge from zero to 120 \(\mu \text{T}\). SVS method achieves an average field strength of \(119,971.54~n\text{T}\) with an RMSE of 698.71 and field fluctuations from \(118,248~n\text{T}\) to \(122,447~n\text{T}\), and it converges to the target field in \(0.48~\text{s}\), demonstrating a fast convergence yet low accuracy and stability in the field generation. ATLMS obtains an average field strength of \(120,012.88~n\text{T}\) with a high RMSE of 671.05, and fluctuations ranging from \(117,923~n\text{T}\) to \(121,628~n\text{T}\), indicating an inaccurate field generation.

The experiment results demonstrate that the proposed convex combination coil control enables the magnetic field generation with rapid response, accurate and stable field generation with fast convergence in a shielded environment. In the next section, we will conduct another experiment in an unshielded environment to check how the designed testbed works with noises in the surrounding environment. 


\subsection{Field generation validation in an unshielded environment}
\label{sec:4.2}

We conduct the field generation in an unshielded environment to test how the designed testbed works with noises from the surrounding environment. In this experiment, we consider the noise in the magnetic field measurement \(\textbf{x}(n)\) and the noise \(\upsilon(n)\) from the sensors and surrounding environment. For both \(\textbf{x}(n)\) and \(\upsilon(n)\), we use the signal-to-noise ratio (SNR) values of \(10~\text{dB}\) and \(30~\text{dB}\), as suggested in \cite{Bershad_2020,Vahidpour_2020}. We compare our coil control with LMS, SVS, and ATLMS, and the parameters of those control approaches are set to their optimal values. The weight update period in the proposed control is set as \(T_o = 2\), and the initial controller weight for the convex combination method is \(\omega=[0.8,0.5]^T\). The full parameter settings for the experiments are shown in \Cref{tab:4}.


\begin{table}[ht]
\setlength\tabcolsep{3pt}
\centering
\caption{The experiment parameter settings}
\label{tab:4}
\begin{tabularx}{0.75\linewidth}{c c c X}
\toprule
\textbf{SNR} & \textbf{\begin{tabular}[c]{@{}c@{}}Input\\ Interference\end{tabular}} & \textbf{Method} & \textbf{Parameters} \\ 
\midrule
\multirow{4}{*}{\(10~\text{dB}\)} & \multirow{2}{*}{\(x(n)\!\!\sim\!\! N(0,1)\)} & LMS & \(\mu {=} 0.01\) \\ 
 &  & SVS & \(\alpha {=} 6\), \(\beta {=} 0.2\) \\ 
 & \multirow{2}{*}{\(v(n) \!\!\sim\!\! N(0,1)\)} & ATLMS & \(\alpha {=} 1000\), \(\beta {=} 0.08\), \(m {=} 1000\), \(n {=} 500\) \\ 
 &  & Convex & \(\alpha {=} 1000\), \(\beta {=} 0.08\), \(\gamma(0) {=} 0.5\), \(\gamma_o {=} 0.55\) \\ 
\midrule
\multirow{4}{*}{\(30~\text{dB}\)} & \multirow{2}{*}{\(x(n) \!\!\sim\!\! N(0,1)\)} & LMS & \(\mu {=} 0.005\) \\ 
 &  & SVS & \(\alpha {=} 4\), \(\beta {=} 0.15\) \\ 
 & \multirow{2}{*}{\(v(n) \!\!\sim\!\! N(0,1)\)} & ATLMS & \(\alpha {=} 500\), \(\beta {=} 0.01\), \(m {=} 900\), \(n {=} 500\) \\ 
 &  & Convex & \(\alpha {=} 500\), \(\beta {=} 0.01\), \(\gamma(0) {=} 0.5\), \(\gamma_o {=} 0.55\) \\ 
\bottomrule
\end{tabularx}
\end{table}

\begin{figure}[htbp]
\centering
\subfigure[]
{
    \begin{minipage}[b]{0.45\linewidth}
        \centering
        \includegraphics[width=\textwidth]{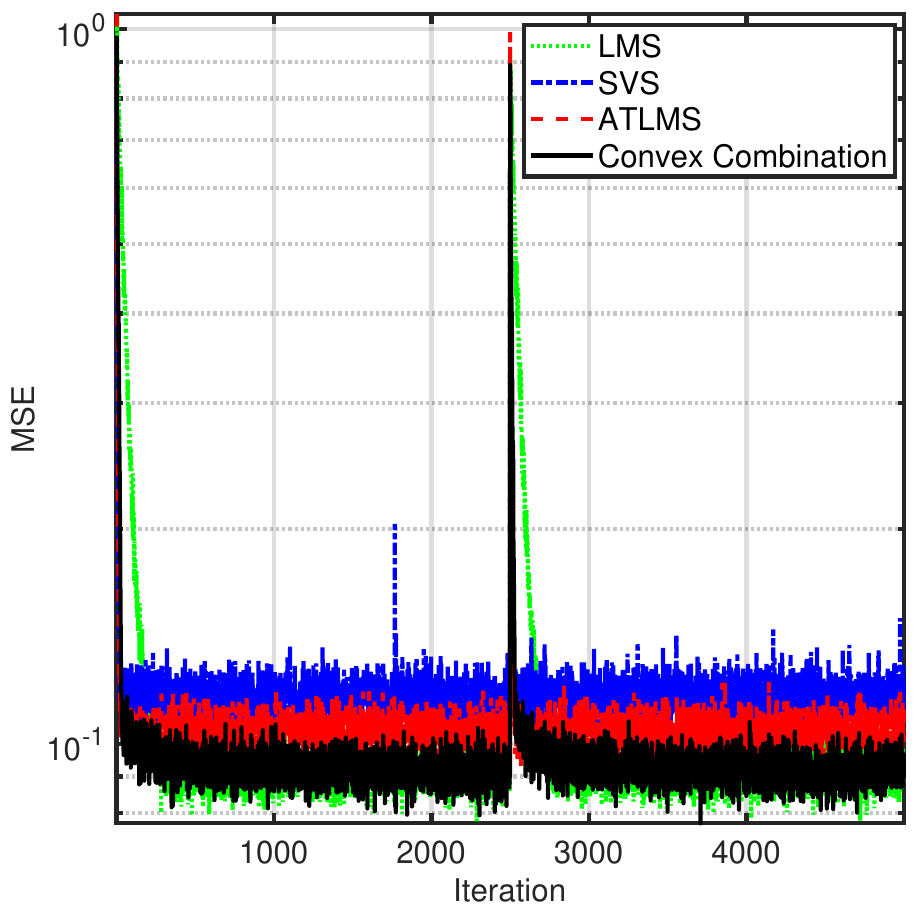}
    \end{minipage}
    \label{fig:12a}
}
\subfigure[]
{
 	\begin{minipage}[b]{0.45\linewidth}
        \centering
        \includegraphics[width=\textwidth]{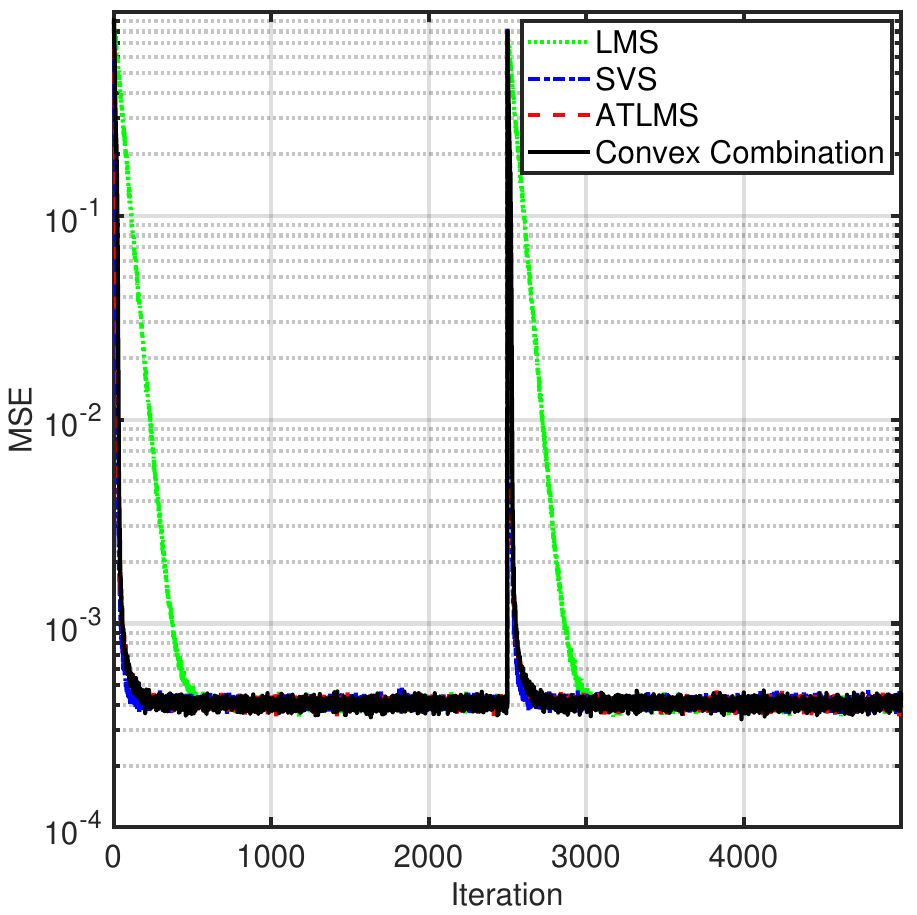}
    \end{minipage}\label{fig:12b}
}
\caption{MSE of the controlled field generation under different SNR: (a) SNR~=~10~\text{dB}, (b) SNR~=~30~\text{dB}.}
\label{fig:12}
\end{figure}

We first conduct the experiment under the low-SNR environment of \(10~\text{dB}\). We visualize the MSE curve for this experiment \Cref{fig:12a}, and detail the numerical results in \Cref{tab:5}. The results indicate that all approaches converge. In comparison, LMS converges more slowly and requires more iterations than the remaining, but it achieves a lower MSE and shows a stable field generation with less fluctuations. SVS provides the fastest convergence but exhibits significant fluctuations with the highest MSE. Compared to SVS, ATLMS improves with less fluctuations and lower MSE. The MSE curve under the proposed convex combination coil control demonstrates better convergence and lower MSE, indicating its advantages in balancing the convergence and stability in achieving accurate field generation. Note that when the noise is reintroduced after stabilization of the field generation at the 2500-th iteration, the proposed coil control can still converge rapidly and maintain stability with accurate field generation.

We then conduct the experiments under a high signal-to-noise ratio of \(30~\text{dB}\). We visualize the experiment results regarding the MSE between the target field strength and the measured strength in \Cref{fig:12b}, and we provide the numerical results in \Cref{tab:5}. We observe from the results that LMS takes 1271 iterations before it converges, which is significantly longer than the remaining approaches. Even though, LMS provide the least MSE thus is the most accurate one in feild generation. Compared to LMS, SVS, and ATLMS, the proposed convex combination method demonstrates better convergence with low MSE without demanding large number of iterations, while maintaining stability when there is inference in the surrounding environment.


\begin{table}[htpb]
\centering
\caption{Numerical results of the coil generation under an unshielded environment.}
\label{tab:5}
\begin{tabularx}{0.75\linewidth}{lcccX}
\toprule
\textbf{SNR} & \textbf{Method} & \textbf{Order} & \textbf{Iteration number} & \textbf{MSE} \\ 
\midrule
\multirow{4}{*}{\(10~\text{dB}\)} & LMS & \(2\) & \(261\) & \(0.0896\) \\ 
 & SVS & \(2\) & \(22\) & \(0.1210\) \\ 
 & ATLMS & \(2\) & \(26\) & \(0.1143\) \\ 
 & Convex & \(2\) & \(33\) & \(0.0904\) \\ 
\midrule
\multirow{4}{*}{\(30~\text{dB}\)} & LMS & \(2\) & \(1,271\) & \(3.881 \times 10^{-4}\) \\ 
 & SVS & \(2\) & \(44\) & \(5.192 \times 10^{-4}\) \\ 
 & ATLMS & \(2\) & \(49\) & \(4.909 \times 10^{-4}\) \\ 
 & Convex & \(2\) & \(55\) & \(3.942 \times 10^{-4}\) \\ 
\bottomrule
\end{tabularx}
\end{table}

\section{Conclusions}
\label{sec:conclusion}
This work designs, optimizes, builds, and validates a hardware-in-the-loop simulation testbed to facilitate geomagnetic navigation experimentation. The designed testbed can rapidly, accurately, and stably generate the magnetic field that meets the geographic navigation requirements. The designed testbed generates a uniform magnetic field by a square Helmholtz coil, where the field generation is determined by our proposed convex combination coil control. Our coil control ensures a balance between convergence and stability of the filed generation toward an accurate magnetic field synthesis. Based on our results, we suggest future works to extend and generate three-dimensional magnetic field. While such a field is fully fledged in supporting geomagnetic navigation, it is challenging to synchronize all the algorithms and hardware for the field generation considering the quality indicators like accuracy, convergence, and stability of the field generation. It will also be interesting to check how the improvement in geomagnetic navigation algorithms can tolerate imperfections in field synthesis in lab conditions.

\bibliographystyle{IEEEtran}
\bibliography{references}

\end{document}